%% file: main.tex
\documentclass[10pt,twocolumn,letterpaper]{article}

\usepackage[pagenumbers]{wacv} 

\usepackage{multirow}
\usepackage{array}
\usepackage{mathtools}
\usepackage{bm}
\usepackage{siunitx}
\usepackage{makecell}
\usepackage{tabularx}
\usepackage{adjustbox}
\usepackage{pifont}
\usepackage{algorithm}
\usepackage[noend]{algpseudocode}
\usepackage[pagebackref,breaklinks,colorlinks,allcolors=blue]{hyperref}
\usepackage[capitalize]{cleveref}

\definecolor{wacvblue}{rgb}{0.21,0.49,0.74}

\def\wacvPaperID{*****} 
\def\confName{WACV}
\def\confYear{2027}

\newcommand{\method}{MP-MNCA\xspace}

\newcommand{\LN}{\operatorname{LN}}

\title{MP-MNCA: Morphology-Prior Masked Neighbor Cross-Attention for Efficient Spatial Domain Identification}

\author{
Sidharthan S C\thanks{Equal contribution.}\\
IIT Madras\\
{\tt\small be21b039@smail.iitm.ac.in}
\and
Sanjeev Manivannan\footnotemark[1]\\
IIT Madras\\
{\tt\small sanjeev.manivannan@gmail.com}
}

\begin{document}
\maketitle

\begin{abstract}
Spatial transcriptomics jointly captures molecular state, tissue morphology, and spatial organization, yet spatial-domain methods are typically compared only at late training checkpoints, obscuring differences in convergence efficiency. We introduce \method, a morphology-prior masked neighbor cross-attention framework that preserves the full 3,000-dimensional highly-variable-gene representation while selectively routing molecular information across spatial neighbors, with histological similarity incorporated as a continuous attention prior under a self-supervised masked neighbor-contrastive objective. Across 12 heterogeneous human dorsolateral prefrontal cortex (DLPFC) sections, \method is evaluated against established spatial-domain baselines using both convergence trajectories and epoch-wise spatial-domain visualizations. Our results show that \method (i) reaches coherent spatial-domain structure in substantially fewer training epochs, as reflected quantitatively by convergence curves and qualitatively by progressive spatial grids, and (ii) maintains strong clustering quality, achieving a mean peak refined ARI of \textbf{0.546} versus \textbf{0.549} for the strongest baseline (STAIG), while attaining the highest mean refined NMI of \textbf{0.657} versus \textbf{0.640}. Moreover, \method outperforms STAIG on \textbf{7 of 12} sections in ARI and \textbf{11 of 12} in NMI, demonstrating faster convergence without sacrificing spatial-domain quality.
\end{abstract}

\section{Introduction}
Spatial transcriptomics (ST) jointly measures transcriptome-wide molecular activity, tissue morphology, and physical position, making it possible to study tissue organization without dissociating cells from their native context  \citep{stahl2016visualization,asp2020spatial,longo2021integrating}. Modern ST platforms provide spatial coordinates with histological images, allowing molecular, spatial, and morphological information to be analyzed jointly \citep{maynard2021transcriptome,moses2022museum}. A central computational problem in ST is therefore \emph{spatial domain identification}: grouping measurement locations into coherent anatomical tissue regions without requiring domain labels during representation learning \citep{hu2021spagcn,zhao2021bayesspace,long2023graphst,yuan2024benchmark}. Such domains capture organized biological structure and form the basis for downstream analysis of tissue architecture, regional gene-expression programs, and disease-associated spatial variation \citep{asp2020spatial,longo2021integrating,moses2022museum}.

ST datasets such as the human dorsolateral prefrontal cortex (DLPFC) dataset can be viewed hierarchically as \emph{donors $\rightarrow$ tissue sections $\rightarrow$ spatial capture spots} \citep{maynard2021transcriptome,pardo2022spatiallibd}. DLPFC contains 12 independently profiled sections from three donors, with four sections per donor \citep{maynard2021transcriptome,pardo2022spatiallibd}. Each section contains approximately 4,000 spatially localized Visium spots, where every spot is associated with a high-dimensional gene-expression profile, a two-dimensional coordinate, and a corresponding H\&E image region \citep{maynard2021transcriptome,shang2022spatial}. For evaluation, spots were retrospectively annotated into cortical layers $\mathrm{L1}$--$\mathrm{L6}$ and white matter (WM) using cytoarchitecture and molecular markers \citep{maynard2021transcriptome}. This donor--section--spot hierarchy is representative of ST datasets in which spatially arranged molecular observations are embedded within a shared histological context \citep{stahl2016visualization,longo2021integrating,moses2022museum}. Despite this structured organization, recovering biologically meaningful spatial domains without using anatomical annotations during training remains difficult due to four major challenges:

The \textit{first major challenge} is inter-donor and inter-section variability under label-free learning. DLPFC provides cortical-layer annotations for retrospective evaluation, but spatial-domain methods must recover these structures without using the labels during representation learning \citep{maynard2021transcriptome,long2023graphst,yuan2024benchmark}. Individual sections differ in sequencing coverage, tissue geometry, molecular composition, and preparation, while additional donor- and batch-specific variation further shifts their distributions \citep{maynard2021transcriptome,long2023graphst,yang2025staig}. Accordingly, methods such as GraphST, MuCoST, and STAIG, discussed in Sec.~\ref{sec:related_work}, learn from unlabeled molecular and spatial structure rather than anatomical supervision \citep{long2023graphst,mucost2024,yang2025staig}. Spatial-domain identification is therefore fundamentally a representation-learning problem: the learned features must remain discriminative across section- and subject-specific variation \citep{long2023graphst,mucost2024,yang2025staig}.

The \textit{second major challenge} is that a Visium spot is a mixed local observation rather than an isolated semantic object. Each spot can contain transcripts from multiple cells and cell types, while its paired histological observation is likewise a local H\&E patch \citep{long2023graphst,yang2025staig}. As a result, spot-level expression can vary substantially even within the same anatomical region because of cellular composition, measurement sparsity, and local biological heterogeneity \citep{long2023graphst,tang2023sigra}. Spatial context is therefore essential: methods that exploit neighboring relationships recover more coherent tissue domains than purely pointwise clustering \citep{hu2021spagcn,long2023graphst,yuan2024benchmark}. This motivates neighborhood-aware representations in which nearby spots contribute contextual information to the spot representation \citep{hu2021spagcn,long2023graphst,mucost2024,yang2025staig}.

The \textit{third major challenge} is that spatial proximity does not imply equal neighbor informativeness. Coordinate-based neighborhoods provide useful candidate interactions, but they can also connect spots across biological boundaries \citep{yang2025staig,long2023graphst}. STAIG addresses such cross-domain edges through morphology-guided edge dropping, while MuCoST supplements spatial adjacency with a separate co-expression graph \citep{yang2025staig,mucost2024}. GraphST aggregates coordinate-defined neighbors with a GCN encoder \citep{long2023graphst}, MuCoST learns from spatial and co-expression graphs with a shared GCN autoencoder \citep{mucost2024}, and STAIG propagates information over morphology-augmented graphs \citep{yang2025staig}. Thus, although these methods adapt neighborhood structure in different ways, information exchange is still primarily through graph construction or propagation rather than an explicit center-conditioned molecular comparison with each neighbor \citep{hu2021spagcn,long2023graphst,mucost2024,yang2025staig}. The key challenge is therefore not only determining \emph{which spots are adjacent}, but \emph{how informative each neighbor is for the specific center spot} \citep{yang2025staig,mucost2024}.

The \textit{fourth major challenge} is that \textit{histological similarity} is informative but not sufficient by itself. For example, DLPFC layers were annotated using both cytoarchitecture and molecular markers (e.g., \emph{MBP}, \emph{PCP4}), rather than H\&E morphology alone \citep{maynard2021transcriptome}. Conversely, gene expressions are also imperfect, with several cortical marker genes exhibiting noisy spot-level expression \citep{tang2023sigra}. Histology therefore provides complementary structural evidence, as demonstrated by morphology-aware methods such as STAIG \citep{yang2025staig}, but image-derived cues can vary with staining and appearance \citep{yang2025staig}. Thus, neither morphology nor gene expression alone is a sufficient semantic signal; the challenge is to use histology as complementary evidence while preserving molecular information as the primary representation of spot identity \citep{maynard2021transcriptome,tang2023sigra,yang2025staig}.

\noindent\textbf{Our Work:} To address these challenges, we introduce \method, a morphology-prior masked
neighbor cross-attention framework that represents each spot as its intrinsic
molecular identity together with a context-dependent neighborhood correction.
We retain the full $G=3000$ HVG representation and use each center spot as a
query over its $k=6$ coordinate-defined spatial neighbors, whose expression profiles provide keys and values for center-specific molecular
compatibility. Histological similarity and relative spatial position
enter the attention logits as soft priors, while masked two-view
self-supervision learns the representation without anatomical domain labels.
The resulting neighborhood context is added residually to the original
spot profile,
\begin{equation}
\hat{\mathbf{x}}_i =
\LN\left(
\mathbf{x}_i + \Delta\mathbf{x}_i^{\mathrm{context}}
\right),
\label{eq:contextual_augmentation}
\end{equation}
which we term \emph{contextual molecular augmentation}. This
preserves original molecular dimensionality while enabling selective,
morphology-guided information exchange between spatial neighbors. Our \textit{main contributions} are:

\noindent$\bullet$ \textbf{Full-dimensional morphology-prior cross-attention.} We introduce a neighbor cross-attention encoder that preserves the 
$3000$-dimensional HVG-representation and learns center-specific
information transfer between spatial neighbors. Histological
similarity and relative position act as soft attention priors instead of
replacing molecular similarity or defining a fixed propagation operator as in Eq.\ref{eq:contextual_augmentation}.

\noindent$\bullet$ \textbf{Masked self-supervised neighborhood learning.}
We train \method independently on each tissue section using two independently
masked gene-expression views and a spatial neighbor-contrastive objective,
without using anatomical domain labels during representation learning. The
resulting embedding preserves intrinsic spot identity through residual
contextual molecular augmentation.

\noindent$\bullet$ \textbf{Convergence-focused spatial-domain evaluation.}
Beyond final-checkpoint performance, we evaluate how spatial-domain structure
emerges throughout optimization using checkpoint-wise ARI/NMI trajectories and
epoch-wise spatial-domain visualizations. Across the 12 DLPFC sections,
\method reaches coherent spatial organization substantially earlier than
established spatial-domain baselines while retaining competitive peak
clustering quality.

\noindent\textbf{Experiments:}
We evaluate \method on all 12 DLPFC sections against SpaGCN, GraphST, MuCoST, and STAIG \citep{hu2021spagcn,long2023graphst,mucost2024,yang2025staig}.
Beyond late-stage performance, we track refined adjusted Rand index (ARI) \citep{hubert1985comparing} and normalized mutual information (NMI) \citep{strehl2002cluster} throughout training and examine epoch-wise spatial domain visualizations to assess both numerical and qualitative convergence. We further report section-wise peak performance and ablations of molecular cross-attention, morphology priors, and positional information. \method reaches competitive spatial-domain structure in substantially fewer epochs, with mean
peak refined ARI of \textbf{0.546} versus \textbf{0.549} for the strongest baseline and the highest mean refined NMI of \textbf{0.657} versus \textbf{0.640}. It further outperforms STAIG on \textbf{7/12} sections in ARI and in \textbf{11/12} sections in NMI, demonstrating faster convergence without sacrificing spatial-domain quality.

\section{Related Work}
\label{sec:related_work}
\paragraph{Graph-based spatial representation learning:}
A common formulation in spatial transcriptomics represents capture spots as graph nodes, associates each node with its gene-expression profile, and uses graph structure to determine which spots exchange information. The principal distinction between existing methods lies in how these inter-spot relationships are constructed and exploited. SpaGCN incorporates spatial coordinates and histological information into its graph and applies graph convolution to aggregate expression from neighboring spots \citep{hu2021spagcn}. GraphST instead constructs a spatial neighborhood graph directly from spot coordinates and employs a GCN encoder with self-supervised contrastive learning to produce spatially informed latent representations \citep{long2023graphst}. MuCoST extends this formulation through multiple graph views: a spatial-adjacency graph captures local proximity, a co-expression graph connects spots with similar molecular profiles, and a shuffled graph provides a repulsive contrastive view; these graphs are processed using a shared GCN autoencoder \citep{mucost2024}. STAIG similarly begins with a coordinate-derived spatial graph in which gene-expression profiles constitute the node information, but uses self-supervised histological embeddings to guide graph augmentation by controlling edge-removal probabilities and filtering potentially homologous negative pairs during contrastive learning \citep{yang2025staig}.

Despite their different graph-construction and training strategies, these approaches principally encode neighborhood information through graph propagation or graph augmentation, typically producing a compressed latent representation of each spot. MP-MNCA retains the useful spatial-neighborhood assumption but changes the mechanism by which molecular information is exchanged. Spatial coordinates define only a small candidate neighborhood; they do not prescribe how strongly every neighbor should contribute. Instead, the center spot queries the full HVG representations of its neighbors through learned cross-attention, producing center-dependent molecular routing. Histology is introduced as a continuous morphological prior on the corresponding attention logits, so that morphological evidence modulates \emph{how much molecular information should be trusted from each neighbor} rather than serving as another feature stream to be fused or merely determining a graph augmentation. Thus, genes carry transmitted content, spatial proximity determines candidate interactions, and morphology modulates reliability of those interactions.

\noindent\textbf{Self-supervised visual features:}
We use frozen BYOL embeddings for morphology. BYOL learns visual representations by matching online and exponentially averaged target networks without explicit negative samples \cite{grill2020byol}. In our framework, these image embeddings are not further trained by the ST objective; only their local similarities enter the attention prior.

\begin{figure*}[t]
    \centering
    \includegraphics[width=\textwidth]{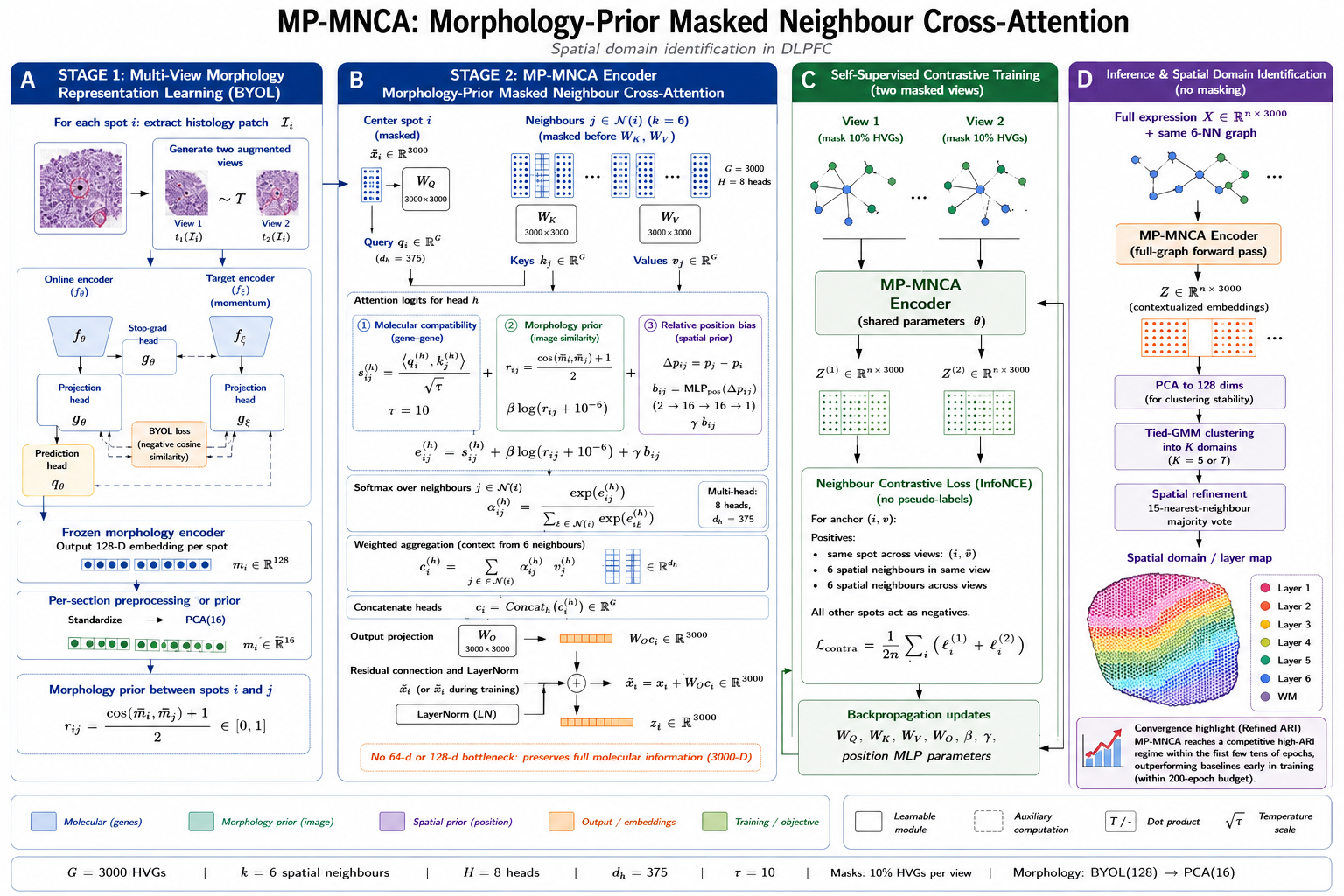}
    \caption{\textbf{MP-MNCA architecture:} (A) Multi-view BYOL learns frozen morphology embeddings used to construct a continuous image-similarity prior. (B) MP-MNCA performs full-dimensional center-to-neighbor molecular cross-attention over six spatial neighbors, combining gene compatibility, morphology prior, and relative-position bias before residual contextualization. (C) Two independently masked molecular views are optimized with a self-supervised neighbor-contrastive loss without pseudo-labels or layer supervision. (D) Inference uses unmasked expression followed by PCA-assisted tied-GMM clustering and spatial refinement.}
    \label{fig:architecture}
\end{figure*}

\section{Method}
\label{sec:method}

\textbf{Overview:} Let a tissue section contain $n$ spatial transcriptomics spots. For spot $i$,
we denote its gene-expression vector, spatial coordinate, and local histology
patch by
\begin{equation}
    \mathbf{x}_i\in\mathbb{R}^{G},\qquad
    \mathbf{p}_i\in\mathbb{R}^{2},\qquad
    \mathcal{I}_i,
\end{equation}
respectively. We use the top $G=3000$ highly variable genes (HVGs) and learn
a full-dimensional representation
\begin{equation}
    f_{\theta}:(X,M,P)\rightarrow Z,
    \qquad
    Z\in\mathbb{R}^{n\times G}.
\end{equation}
Our framework consists of two stages. First, a \textit{multi-view BYOL encoder} learns
a morphology representation from the histology surrounding each spot.
Second, \textit{\method} uses the resulting morphology similarity as a soft prior for
center-conditioned molecular cross-attention over spatial neighbors. The
morphology representation controls \emph{where molecular information is
routed}; the transmitted content itself always comes from gene expression.


\subsection{Stage I: Multi-View Morphology Encoding}
\label{sec:method_morphology}

For each spot $i$, we extract a local histological patch $\mathcal{I}_i$
centered at its spatial coordinate. To learn morphology feature \textbf{\textit{representation}} that are
invariant to nuisance appearance variation, we employ a BYOL encoder with
multiple stochastic views of the same patch. Specifically, two transformations
$t_1,t_2\sim\mathcal{T}$ produce
\begin{equation}
    \mathcal{I}_i^{(1)}=t_1(\mathcal{I}_i),
    \qquad
    \mathcal{I}_i^{(2)}=t_2(\mathcal{I}_i).
\end{equation}
The online and target BYOL branches are optimized to produce consistent
representations for these two views without requiring tissue annotations.
After pretraining, the image encoder is frozen and a $d_m=128$ dimensional embedding $\mathbf{m}_i\in\mathbb{R}^{128}$
is extracted for every spot.

The morphology encoder is not jointly optimized with the molecular model. Instead, image embeddings are standardized independently within each tissue section and reduced to 16 dimensions using Principal Component Analysis (PCA) \cite{pearson1901pca} solely for computing the morphology prior:
\begin{equation}
    \bar{M}
    =
    \operatorname{PCA}_{16}
    \left(
        \operatorname{Standardize}(M)
    \right)
\end{equation}
Importantly, image embeddings are never concatenated with gene expression and
are never used as molecular value vectors. For a center spot $i$ and neighbor $j$, we define their morphology agreement
using cosine similarity,
\begin{equation}
    s_{ij}^{\mathrm{img}}
    =
    \frac{
        \langle\bar{\mathbf m}_i,\bar{\mathbf m}_j\rangle
    }{
        \|\bar{\mathbf m}_i\|_2
        \|\bar{\mathbf m}_j\|_2
    },
    \qquad
    r_{ij}
    =
    \frac{1+s_{ij}^{\mathrm{img}}}{2}
    \in[0,1].
\label{eq:morph_prior}
\end{equation}
Thus, morphology provides a continuous measure of whether a neighboring spot
is visually compatible with the center rather than defining a hard
image-derived graph or cluster assignment.

\subsection{Stage II: Morphology-Prior Masked Neighbor Cross-Attention}
\label{sec:method_mpmnca}

\paragraph{Spatial candidate neighborhood:}
For each center spot $i$, we define a local seven-spot receptive field
consisting of the center and its six nearest spatial neighbors. Let
\begin{equation}
    \mathcal{N}(i)
    =
    \underset{j\neq i}{\operatorname{kNN}}_{6}
    (\mathbf{p}_i),
    \qquad
    \mathcal{S}(i)=\{i\}\cup\mathcal{N}(i),
\label{eq:knn}
\end{equation}
where $|\mathcal{N}(i)|=6$ and $|\mathcal{S}(i)|=7$.
Consequently, each local receptive field contains seven molecular profiles:
the center $\mathbf{x}_i$ and six neighboring profiles
$\{\mathbf{x}_j:j\in\mathcal{N}(i)\}$.
The coordinate graph only determines which spots are eligible to exchange
information, not their aggregation weights.

\paragraph{Masked molecular views:}
To learn representations robust to partial molecular corruption, two
independent views of the complete HVG matrix are generated during training.
For $v\in\{1,2\}$,
\begin{equation}
    \tilde{\mathbf{x}}_{\ell}^{(v)}
    =
    \mathbf{m}^{(v)}\odot\mathbf{x}_{\ell},
\end{equation}
where $\mathbf{m}^{(v)}\in\{0,1\}^{G}$ masks approximately $10\%$ of gene
dimensions. The selected gene columns are shared across all spots within a
view, while the two views are masked independently. Spatial neighborhoods and
image-derived morphology priors remain unchanged.

\paragraph{Center-conditioned molecular cross-attention:}
For each center spot, the center molecular representation generates the query,
while its six spatial neighbors generate keys and values:
\begin{equation}
\begin{aligned}
    \mathbf{q}_i = W_Q\tilde{\mathbf{x}}_i,\quad
    \mathbf{k}_j = W_K\tilde{\mathbf{x}}_j,\quad
    \mathbf{v}_j = W_V\tilde{\mathbf{x}}_j,
\end{aligned}
\label{eq:qkv}
\end{equation}
where $W_Q,W_K,W_V\in\mathbb{R}^{G\times G}$ and $j\in\mathcal{N}(i)$.
The model therefore retains the full $G=3000$ dimensional molecular state
rather than introducing a learned low-dimensional bottleneck. 

\noindent We use $H=8$ attention heads, giving $d_h=\frac{G}{H}=375$.
For head $h$, molecular compatibility between center $i$ and neighbor $j$ is
\begin{equation}
    s_{ij}^{\mathrm{gene},(h)}
    =
    \frac{
        \left\langle
            \mathbf{q}_i^{(h)},
            \mathbf{k}_j^{(h)}
        \right\rangle
    }{\sqrt{\tau}},
\label{eq:gene_score}
\end{equation}
where $\tau=10$ is the configured attention scale.

\paragraph{Morphology-prior routing:}
Molecular compatibility alone may assign high attention to a transcriptionally
similar neighbor whose local tissue morphology is inconsistent with the center.
We therefore inject the image-derived similarity from
Eq.~\eqref{eq:morph_prior} directly into attention logits. Spatial orientation is further encoded using the relative coordinate
\begin{equation}
    \Delta\mathbf{p}_{ij}
    =
    \mathbf{p}_j-\mathbf{p}_i,
\end{equation}
and a positional network $\operatorname{MLP}_{\mathrm{pos}}:
    \mathbb{R}^{2}\rightarrow
    \mathbb{R}^{16}\rightarrow
    \mathbb{R},$
\begin{equation}
    b_{ij}
    =
    \operatorname{MLP}_{\mathrm{pos}}
    (\Delta\mathbf{p}_{ij}),
\label{eq:position}
\end{equation}
with GELU non linearity. The resulting scalar $b_{ij}$
allows the attention mechanism to learn direction- and displacement-dependent spatial preferences rather than treating all six neighboring positions equivalently. The complete attention energy for head $h$ is
\begin{equation}
    e_{ij}^{(h)}
    =
    \underbrace{
    \frac{
        \langle
            \mathbf{q}_i^{(h)},
            \mathbf{k}_j^{(h)}
        \rangle
    }{\sqrt{\tau}}
    }_{\text{molecular compatibility}}
    +
    \underbrace{
        \beta\log(r_{ij}+\epsilon)
    }_{\text{morphology prior}}
    +
    \underbrace{
        \gamma b_{ij}
    }_{\text{pos. bias}}
\label{eq:attention_energy}
\end{equation}
with $\epsilon=10^{-6}$. The scalars $\beta$ and $\gamma$ are learnable and are initialized to $1.0$ and $0.5$, respectively. Attention is normalized only over the six eligible neighbors,
\begin{equation}
    \alpha_{ij}^{(h)}
    =
    \frac{
        \exp(e_{ij}^{(h)})
    }{
        \displaystyle
        \sum_{\ell\in\mathcal{N}(i)}
        \exp(e_{i\ell}^{(h)})
    }.
\label{eq:attention}
\end{equation}
Equivalently, $\alpha_{ij}^{(h)}
    \propto
    \exp\!\left(
        s_{ij}^{\mathrm{gene},(h)}
    \right)
    r_{ij}^{\beta}
    \exp(\gamma b_{ij})$,
showing that morphology acts as a multiplicative routing prior. It changes the
relative probability that a neighboring molecular message is transmitted,
without replacing that message with image information.

\paragraph{Contextual molecular augmentation:}
Each attention head computes a center-specific molecular context
\begin{equation}
    \mathbf{\Delta_i}^{(h)} \equiv \mathbf{c}_i^{(h)}
    =
    \sum_{j\in\mathcal{N}(i)}
        \alpha_{ij}^{(h)}
        \mathbf{v}_j^{(h)}
    \in\mathbb{R}^{d_h}.
\label{eq:head_context}
\end{equation}
The head-wise contexts are concatenated and projected back to the original HVG
space:
\begin{equation}
    \mathbf{c}_i
    =
    W_O
    \operatorname{Concat}_{h=1}^{H}
    \left(
        \mathbf{c}_i^{(h)}
    \right),
    W_O\in\mathbb{R}^{G\times G}.
\label{eq:context}
\end{equation}

Crucially, the neighboring context \emph{augments} the center spot rather than replacing it. The \textbf{\textit{final representation}} is
\begin{equation}
\boxed{
    \mathbf{z}_i
    =
    \operatorname{LN}
    \left(
        \mathbf{x}_i+\mathbf{c}_i
    \right)
}
\label{eq:residual}
\end{equation}
for an unmasked input. During masked-view training, $\mathbf{x}_i$ in
Eq.~\eqref{eq:residual} is replaced by the respective masked center view.

\paragraph{Self-supervised neighbor contrastive learning:}
The same \method parameters process the two independently masked molecular
views, producing
\begin{equation}
    Z^{(1)},Z^{(2)}
    \in\mathbb{R}^{n\times G}.
\end{equation}
Let
\begin{equation}
    \bar{\mathbf z}_i^{(v)}
    =
    \frac{
        \mathbf z_i^{(v)}
    }{
        \|\mathbf z_i^{(v)}\|_2
    }
\end{equation}
denote the normalized representation. For an anchor $(i,v)$, the positive set contains
\begin{equation}
\mathcal{P}_i^{(v)}
=
\left\{
    (i,\bar v)
\right\}
\cup
\left\{
    (j,v):j\in\mathcal{N}(i)
\right\}
\cup
\left\{
    (j,\bar v):j\in\mathcal{N}(i)
\right\},
\label{eq:positive_set}
\end{equation}
where $\bar v$ denotes the opposite masked view. Thus, each anchor has the same
spot in the other view together with its six spatial neighbors in both views
as positives.

For any pair of normalized embeddings, define
\begin{equation}
    \operatorname{sim}
    \left(
        \bar{\mathbf z}_i^{(v)},
        \bar{\mathbf z}_j^{(u)}
    \right)
    =
    \exp
    \left(
        \frac{
            \left\langle
                \bar{\mathbf z}_i^{(v)},
                \bar{\mathbf z}_j^{(u)}
            \right\rangle
        }{\tau_c}
    \right),
\end{equation}
where $\tau_c$ is the contrastive temperature. Let $\mathcal{A}_i^{(v)}$ contain all valid non-anchor embeddings from both
views. The directional multi-positive contrastive loss is
\begin{equation}
    \ell_i^{(v)}
    =
    -\log
    \frac{
        \displaystyle
        \sum_{(j,u)\in\mathcal{P}_i^{(v)}}
        \operatorname{sim}
        \left(
            \bar{\mathbf z}_i^{(v)},
            \bar{\mathbf z}_j^{(u)}
        \right)
    }{
        \displaystyle
        \sum_{(j,u)\in\mathcal{A}_i^{(v)}}
        \operatorname{sim}
        \left(
            \bar{\mathbf z}_i^{(v)},
            \bar{\mathbf z}_j^{(u)}
        \right)
    }.
\label{eq:contrastive_direction}
\end{equation}
The final objective is symmetric across the two views,
\begin{equation}
    \mathcal{L}_{\mathrm{contra}}
    =
    \frac{1}{2n}
    \sum_{i=1}^{n}
    \left(
        \ell_i^{(1)}
        +
        \ell_i^{(2)}
    \right).
\label{eq:contrastive}
\end{equation}

All remaining spot representations contribute through the contrastive
denominator. In particular, \textbf{no image KMeans assignments, gene
pseudo-labels, anatomical labels, or pseudo-label-based negative masking are
used during \method training}. Backpropagation therefore learns
$W_Q,W_K,W_V,W_O$, the positional MLP, and the routing calibration parameters
$\beta$ and $\gamma$ purely from masked-view consistency, spatial locality,
gene-expression compatibility, and the continuous morphology prior.

\begin{table*}[t]
\centering
\caption{
\textbf{Peak-checkpoint spatial-domain identification on the 12 DLPFC sections.}
We report the maximum refined ARI observed within the same 200-epoch optimization budget and the corresponding refined NMI at that checkpoint, with the selected epoch shown in parentheses. This complementary analysis assesses whether MP-MNCA's faster convergence is achieved without sacrificing peak clustering quality. Higher is better, and the best result for each section and metric is shown in bold.
}
\label{tab:dlpfc_peak_comparison}

\scriptsize
\setlength{\tabcolsep}{3.2pt}
\renewcommand{\arraystretch}{0.9}

\resizebox{\textwidth}{!}{%
\begin{tabular}{l|cc|cc|cc|cc}
\toprule
\multirow{2}{*}{\textbf{Section}}
& \multicolumn{2}{c|}{\textbf{STAIG}}
& \multicolumn{2}{c|}{\textbf{GraphST}}
& \multicolumn{2}{c|}{\textbf{MuCoST}}
& \multicolumn{2}{c}{\textbf{MP-MNCA}} \\

\cmidrule(lr){2-3}
\cmidrule(lr){4-5}
\cmidrule(lr){6-7}
\cmidrule(lr){8-9}

& ARI (ep.) $\uparrow$ & NMI $\uparrow$
& ARI (ep.) $\uparrow$ & NMI $\uparrow$
& ARI (ep.) $\uparrow$ & NMI $\uparrow$
& ARI (ep.) $\uparrow$ & NMI $\uparrow$ \\

\midrule

151507
& \textbf{0.5786 (90)} & 0.6458
& 0.4848 (200) & 0.6507
& 0.5556 (100) & 0.6638
& 0.5734 (15) & \textbf{0.6884} \\

151508
& 0.4938 (200) & 0.6334
& 0.5133 (1) & 0.5565
& 0.4154 (115) & 0.5341
& \textbf{0.5459 (195)} & \textbf{0.6894} \\

151509
& \textbf{0.6310 (160)} & 0.6548
& 0.4915 (195) & 0.5406
& 0.5799 (120) & \textbf{0.6893}
& 0.5420 (50) & 0.6803 \\

151510
& 0.5702 (125) & 0.6525
& 0.4185 (155) & 0.5853
& 0.5480 (110) & 0.6499
& \textbf{0.5704 (5)} & \textbf{0.6600} \\

151669
& 0.4444 (200) & 0.5008
& 0.4417 (190) & 0.4967
& \textbf{0.4908 (185)} & \textbf{0.5494}
& 0.4549 (10) & 0.5038 \\

151670
& 0.5488 (190) & \textbf{0.5974}
& 0.2498 (130) & 0.4209
& \textbf{0.5803 (20)} & 0.5239
& 0.4221 (10) & 0.4473 \\

151671
& 0.5953 (45) & 0.6864
& \textbf{0.8064 (200)} & \textbf{0.7332}
& 0.7464 (195) & 0.7232
& 0.6054 (15) & 0.7207 \\

151672
& 0.5762 (40) & 0.6574
& 0.6756 (10) & 0.6352
& \textbf{0.7151 (180)} & 0.7145
& 0.6064 (150) & \textbf{0.7245} \\

151673
& \textbf{0.5556 (40)} & 0.6556
& 0.3673 (1) & 0.5392
& 0.4791 (50) & 0.5901
& 0.5287 (45) & \textbf{0.6898} \\

151674
& 0.5248 (115) & 0.6503
& 0.4284 (15) & 0.5412
& 0.3442 (1) & 0.5205
& \textbf{0.5663 (90)} & \textbf{0.6902} \\

151675
& 0.5151 (185) & 0.6715
& 0.3700 (5) & 0.5504
& 0.3216 (5) & 0.5142
& \textbf{0.5989 (40)} & \textbf{0.7100} \\

151676
& \textbf{0.5587 (165)} & 0.6686
& 0.3822 (5) & 0.5075
& 0.2133 (50) & 0.3974
& 0.5312 (85) & \textbf{0.6730} \\

\midrule

\textbf{Mean $\pm$ SD}
& \textbf{0.5494 $\pm$ 0.0472}
& 0.6395 $\pm$ 0.0468
& 0.4691 $\pm$ 0.1414
& 0.5631 $\pm$ 0.0773
& 0.4991 $\pm$ 0.1503
& 0.5892 $\pm$ 0.0953
& 0.5455 $\pm$ 0.0547
& \textbf{0.6565 $\pm$ 0.0836} \\

\textbf{Median [IQR]}
& \textbf{0.5572 [0.0544]}
& 0.6537 [0.0175]
& 0.4351 [0.1178]
& 0.5458 [0.0665]
& 0.5194 [0.1824]
& 0.5698 [0.1471]
& 0.5561 [0.0492]
& \textbf{0.6889 [0.0254]} \\

\bottomrule
\end{tabular}%
}
\end{table*}

\section{Experiments}

\subsection{Dataset and Evaluation Protocol}
We evaluate on the 12 human dorsolateral prefrontal cortex (DLPFC) sections introduced by Maynard \etal \cite{maynard2021transcriptome}. The sections span three donors and contain approximately 3,500--4,800 spatial spots each, with annotated cortical layers and white matter providing the reference spatial domains. Performance is evaluated using Adjusted Rand Index (ARI) and Normalized Mutual Information (NMI).

Our evaluation is designed primarily to compare the \emph{convergence behavior of the model architectures}. To minimize differences arising from initialization, all methods use the same section-wise preprocessing, spatial neighborhood construction, cluster count, and common pre-model initialization protocol. Thus, for a given section, each method starts from the same initialized molecular/spatial state before being passed through its respective architecture; only the subsequent representation-learning and message-routing mechanisms differ between models. Architecture-specific trainable parameters necessarily retain their respective parameterizations, but are initialized under the same deterministic experimental protocol.

Following this common starting point, \method, STAIG, GraphST, and MuCoST are optimized and evaluated over the same 200-epoch budget. Refined ARI is measured throughout training and aggregated across the 12 sections to characterize how rapidly each architecture develops spatially discriminative representations and whether this quality is maintained as optimization proceeds. We additionally report the best ARI with NMI at the corresponding checkpoint, to compare attainable clustering quality alongside convergence speed.

\subsection{Implementation Details}
Gene expression is represented using 3,000 log1p-normalized HVGs with six coordinate-defined neighbors per spot. Morphology is encoded using frozen 128-dimensional BYOL features, standardized per section and reduced to 16 dimensions by PCA. \method uses eight attention heads with $d_k=375$, preserving the full 3,000-dimensional molecular representation. During training, 10\% of HVGs are independently masked in each view before the $Q/K/V$ projections; $\beta$ and $\gamma$ are initialized to 1 and 0.5, respectively. 

We use attention dropout 0.1 and Adam with learning rate $3\times10^{-4}$ and weight decay $10^{-5}$. For convergence evaluation, \method, STAIG, GraphST, and MuCoST are trained for 200 epochs under the matched protocol described above, while retaining method-specific repository-supported optimization settings. We compare epoch-wise convergence rather than wall-clock runtime, as per-epoch computational cost differs across architectures.
\subsection{Ablation Protocol}
The ablation study isolates the principal components of the proposed routing mechanism. Removing the morphology prior ($\beta=0$) tests whether histological similarity contributes beyond molecular compatibility, while removing the position term ($\gamma=0$) evaluates the additional role of relative spatial geometry. Setting $\beta=\gamma=0$ isolates molecular cross-attention, and replacing learned attention coefficients with uniform $1/k$ weights tests adaptive routing against fixed neighborhood averaging. Low-dimensional variants further test whether the full-HVG representation is important for the observed behavior.

\section{Results}

\subsection{Convergence Across DLPFC Sections}

\begin{figure}[t]
\centering
\includegraphics[width=\linewidth]{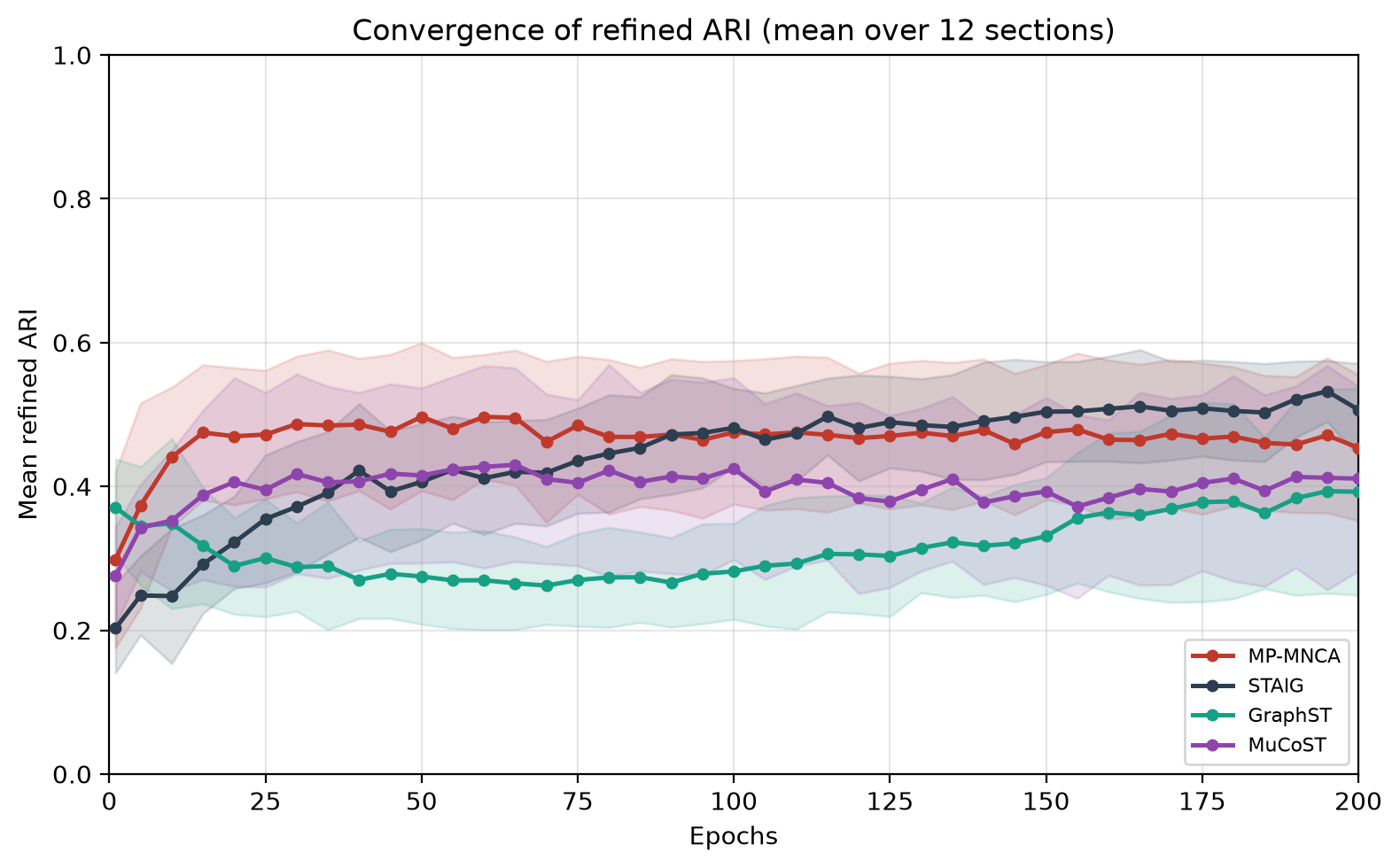}
\caption{
\textbf{Convergence of refined ARI across the 12 DLPFC sections.}
Mean refined ARI over 200 epochs with $\pm$ one standard deviation across sections. \method reaches a competitive regime substantially earlier than the baselines.
}
\label{fig:convergence_mean}
\end{figure}

\Cref{fig:convergence_mean} shows the evolution of refined ARI averaged over all 12 DLPFC sections. \method reaches a competitive high-ARI regime within the first 10--20 epochs and remains comparatively stable thereafter, whereas the baselines require substantially longer optimization or stabilize at lower mean performance. Although STAIG eventually becomes comparable and occasionally exceeds \method at later checkpoints, \method reaches similar clustering quality much earlier, highlighting its stronger convergence behavior. The overlap between the section-wise variance bands at later epochs further motivates evaluating peak attainable quality separately rather than claiming uniform dominance throughout optimization.

\subsection{Peak Attainable Clustering Quality}
\Cref{tab:dlpfc_peak_comparison} shows that faster convergence does not cost peak quality. Within the same 200-epoch budget, \method attains a mean peak ARI of $0.5455\pm0.0547$, essentially matching STAIG ($0.5494\pm0.0472$) and exceeding GraphST and MuCoST, while achieving the highest mean NMI of $0.6565\pm0.0836$ versus $0.6395\pm0.0468$ for STAIG, with the best NMI on 8 of 12 sections. Peak-checkpoint locations reinforce the trend: \method peaks by epoch 50 on 8 of 12 sections against 3 for STAIG, with median peak epochs of 42.5 and 142.5.

\subsection{Qualitative Analysis and Ablations}
\begin{figure*}[t]
\centering
\includegraphics[width=0.92\textwidth]{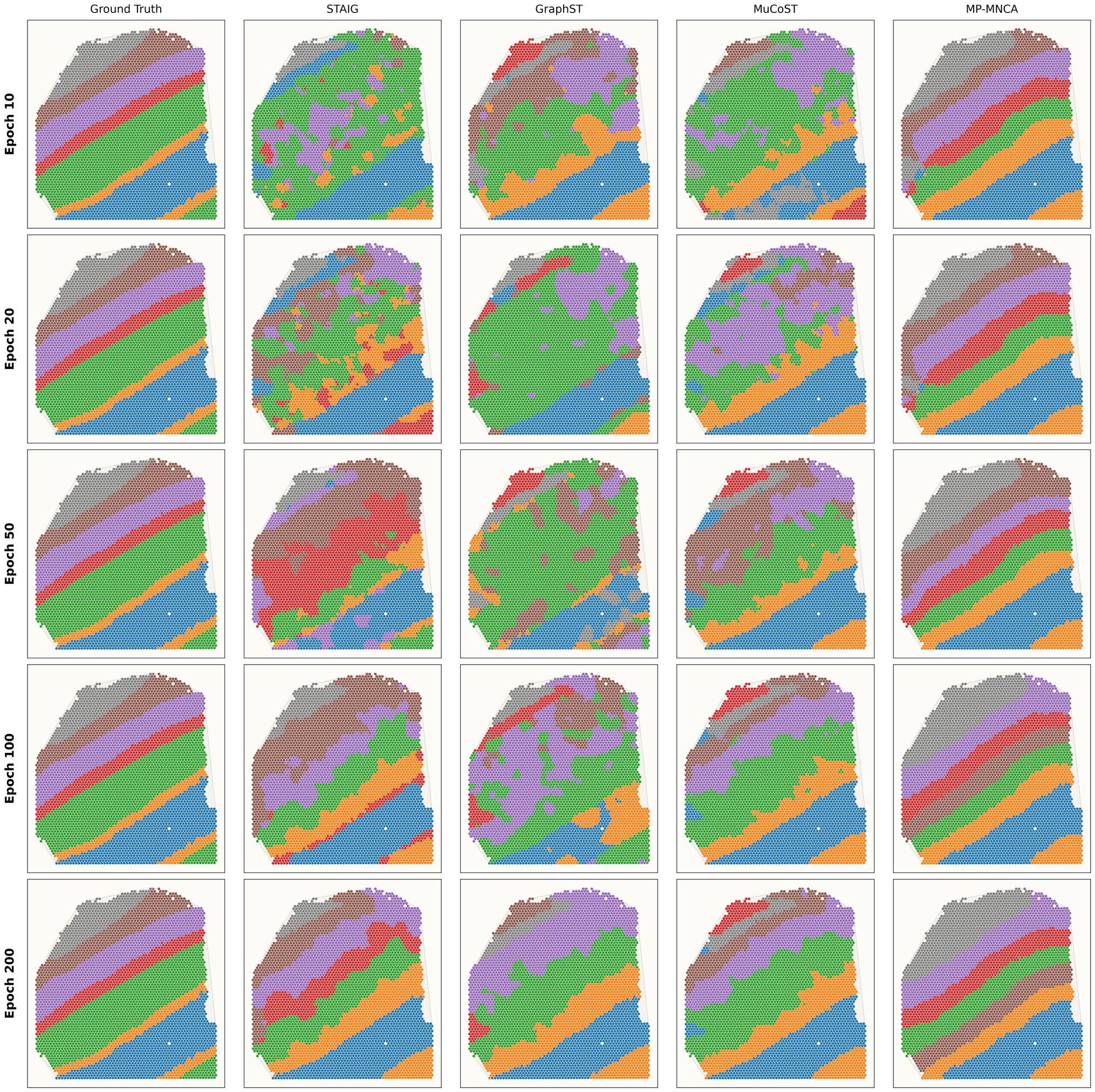}
\caption{
\textbf{Qualitative convergence on DLPFC section 151507.}
Spatial-domain predictions at epochs 10, 20, 50, 100, and 200 for STAIG, GraphST, MuCoST, and \method. \method recovers coherent laminar structure earlier, while competing methods show slower organization or greater domain mixing. Ground truth is shown for reference; evaluation uses permutation-invariant ARI and NMI.
}
\label{fig:convergence_grid_151507}
\end{figure*}

The same dynamics are visible in the predicted partitions. \Cref{fig:convergence_grid_151507} shows section 151507 at epochs 10, 20, 50, 100, and 200: \method recovers the dominant laminar organization within 10--20 epochs and keeps coherent boundaries thereafter, whereas STAIG shows substantial early local mixing before organizing, and GraphST and MuCoST show greater boundary mixing or deformation at intermediate checkpoints. \method peaks at ARI 0.5734 (epoch 15) against 0.5786 for STAIG (epoch 90), with GraphST and MuCoST peaking at epochs 200 and 100, and attains the highest NMI on this section (0.6884).

Ablations support the routing design. Compressing the molecular representation to 128 dimensions drops mean refined ARI to 0.272, 0.243, and 0.234 for the contrastive, gene-cosine, and BYOL variants, confirming the value of the full HVG representation. The $\beta=0$, $\gamma=0$, $\beta=\gamma=0$, and uniform-neighbor controls isolate the contributions of morphology, position, molecular compatibility, and adaptive neighbor weighting under the same six-neighbor support and evaluation pipeline.

\subsection{Attention Routing and Early Convergence}

The observed convergence profile is consistent with the inductive structure of the attention block:
\begin{equation}
    \alpha_{ij}\propto
    \exp\!\left(
    \frac{q_i^\top k_j}{\sqrt{d}}
    +\beta\,m_{ij}
    +\gamma\,p_{ij}
    \right),
\end{equation}
where $j\in\mathcal{N}_6(i)$, and the six-neighbor graph only defines candidate support, while molecular compatibility, morphology, and relative position determine routing strength. This selects informative local context directly in the full HVG space, while the residual preserves the center representation. The resulting inductive bias is consistent with the rapid early ARI increase in \Cref{fig:convergence_mean}, though causal attribution requires component-wise convergence ablations.

\section{Discussion and Conclusion}
\method preserves the full molecular state and learns neighbor-specific routing before compression; the degradation of the 128-dimensional controls shows the gains are not from low-dimensional fusion or spatial smoothing alone. The architecture separates topology, prior, and content: coordinates define candidate neighbors, molecular compatibility sets contribution, and histology modulates reliability through an additive prior that acts multiplicatively after exponentiation-discounting an unreliable molecular message without fusing morphology into the representation. This matters most near tissue boundaries, where morphology and molecular identity diverge.

Evaluation is restricted to DLPFC, but the benchmark spans multiple sections and heterogeneous subjects, and the consistency across sections suggests the gains are not driven by a single tissue instance; additional datasets would nevertheless establish cross-dataset generality. The four $3000\times3000$ projections introduce roughly 36M parameters, motivating low-rank or shared projections.

Because morphology only reweights message passing, the two modalities never share a coordinate system. We therefore also investigated \textbf{SPARC-align}, in which concept-aligned sparse autoencoders~\cite{sparc2026} encode expression and histology into a shared dictionary with consistent per-dimension meaning across modalities, passing the resulting concept codes to the cross-attention stage of Sec.~\ref{sec:method}. Two adaptations proved important: smoothing each spot over its histology-weighted neighborhood before encoding, since sparse autoencoders assume i.i.d.\ samples while capture spots are spatially organized; and giving each modality private dictionary capacity alongside the shared block, since expression and morphology are asymmetrically predictive. With both, SPARC-align approaches established baselines while yielding an interpretable cross-modal concept space that none of the compared methods provide, making concept-aligned sparse coding a promising direction for multimodal spatial transcriptomics. Full formulation, per-section results, convergence, ablations, and limitations appear in the supplementary material.

\bibliographystyle{plainnat}
\bibliography{main}

\input{supplementary}

\end{document}

%% file: supplementary.tex
%
\onecolumn
\noindent
This supplementary includes \textbf{SPARC-align}, the concept-aligned cross-modal
sparse-coding variant referenced. We describe the motivation
(Sec.~\ref{sec:motivation}), the method (Sec.~\ref{sec:method}), results on all
twelve DLPFC sections (Sec.~\ref{sec:results}), and the limitations that determine
where we would take the approach next (Sec.~\ref{sec:discussion}). All numbers are
produced under the evaluation protocol of the main paper, and all training is fully
unsupervised.

\section{From MP-MNCA to SPARC-align}
\label{sec:motivation}

MP-MNCA routes molecular information across spatial neighbors using histological
similarity as a continuous attention prior. Morphology therefore enters as a
\emph{weighting} on message passing: it decides how strongly a neighbor contributes,
but the representation being learned remains a refinement of gene expression. The two
modalities are never placed in a common coordinate system, so one cannot ask which
morphological structure a given molecular program corresponds to.

That question motivates a different formulation. If gene expression and histology were
encoded into a \emph{shared} latent space in which an individual dimension carried the
same meaning for both, the resulting code would be interpretable across modalities
rather than merely informed by one of them. Sparse autoencoders let us do the same via dictionary learning: SPARC~\cite{sparc2026}  learns per-stream encoders into a common dictionary
and enforces alignment by selecting a single shared set of active dimensions from the
summed pre-activations, so both streams must explain a sample using the same concepts.

We therefore built \textbf{SPARC-align}: a two-phase pipeline in which a
concept-aligned sparse autoencoder replaces MP-MNCA's implicit fusion, followed by
MP-MNCA's neighbor cross-attention operating on the resulting concept codes. Applying
SPARC to spatial transcriptomics is not, however, a direct transfer. SPARC assumes
i.i.d.\ samples, whereas capture spots are spatially organized: in DLPFC, a spot and
its six nearest neighbors share a cortical layer $92.3\%$ of the time. Sec.~\ref{sec:method}
describes two modifications that follow from this---spatially smoothing the inputs
before encoding, and relaxing the shared support so each modality retains private
dictionary capacity. These were the only changes, among many evaluated, that improved
clustering.

\section{Method}
\label{sec:method}

\subsection{Streams and notation}

Each section is processed independently. For section with $n$ spots we take two frozen
streams: gene expression $\mathbf{x}^{g}_i \in \mathbb{R}^{3000}$ (the same
highly-variable-gene matrix used by MP-MNCA) and histology
$\mathbf{x}^{m}_i \in \mathbb{R}^{128}$ (the frozen BYOL embedding of the main paper).
Both are standardized per feature within a section; this is required because the two
streams differ in scale by an order of magnitude, and an unstandardized shared
selection would be decided by histology alone. Let $\mathcal{N}(i)$ denote the six
spatial nearest neighbors of spot $i$.

\subsection{Global TopK concept alignment}

Each stream $s \in \{g, m\}$ has an affine encoder into a common $L$-dimensional
dictionary, producing pre-activations
\begin{equation}
\mathbf{h}^{s}_i = W^{s}_{E}\bigl(\mathbf{x}^{s}_i - \mathbf{b}^{s}_{\mathrm{pre}}\bigr) + \mathbf{b}^{s}_{\mathrm{lat}} .
\end{equation}
Alignment is enforced structurally. Rather than sparsifying each stream separately,
the active set is chosen \emph{once} from the summed pre-activations,
\begin{equation}
\mathcal{I}_i = \operatorname{TopK}\Bigl(\textstyle\sum_{s} \mathbf{h}^{s}_i,\; k\Bigr),
\qquad
\mathbf{z}^{s}_i = \mathrm{ReLU}\bigl(\mathrm{gather}(\mathbf{h}^{s}_i, \mathcal{I}_i)\bigr),
\end{equation}
so both streams activate identical dimensions for a given spot while retaining
stream-specific magnitudes. Per-stream affine decoders reconstruct each input, and the
objective combines self- and cross-reconstruction in normalized MSE,
\begin{equation}
\mathcal{L} = \sum_{s} \mathcal{L}_{\mathrm{NMSE}}\bigl(\mathbf{x}^{s}, D_{s}(\mathbf{z}^{s})\bigr)
            + \lambda \sum_{s \neq t} \mathcal{L}_{\mathrm{NMSE}}\bigl(\mathbf{x}^{t}, D_{t}(\mathbf{z}^{s})\bigr),
\end{equation}
with an auxiliary term reviving dormant dictionary atoms. Cross-reconstruction is what
makes the shared support \emph{semantically} aligned rather than merely coincident.

\subsection{Spatial \emph{k}-NN smoothing before alignment}

SPARC treats samples independently, which discards the strongest prior available in
spatial transcriptomics. We inject it \emph{before} encoding, replacing each spot's
features with a morphology-weighted blend of itself and its neighborhood:
\begin{equation}
\tilde{\mathbf{x}}^{s}_i = (1-\alpha)\,\mathbf{x}^{s}_i
  + \alpha \sum_{j \in \mathcal{N}(i)} w_{ij}\, \mathbf{x}^{s}_j ,
\label{eq:smooth}
\end{equation}
\begin{equation}
w_{ij} = \operatorname{softmax}_{j}\bigl(\beta \log s_{ij}\bigr),
\qquad
s_{ij} = \tfrac{1}{2}\bigl(\cos(\mathbf{m}_i, \mathbf{m}_j) + 1\bigr),
\end{equation}
where $s_{ij}$ is the histological similarity prior of the main paper. Weighting by
$s_{ij}$ rather than uniformly means morphologically dissimilar neighbors contribute
less, so laminar boundaries blur less than under a plain neighborhood mean. We use
$\alpha = 0.5$, giving the spot and its neighborhood equal weight; this value was set
by analogy to a normalized adjacency with self-loops and was \emph{not} tuned.

The design rationale is asymmetry between the streams. Histology reconstructs well
(self-NMSE $\approx 0.28$) whereas expression does not ($\approx 0.88$), so the shared
support is dominated by histology and the concept space carries little laminar
information. Smoothing raises the signal-to-noise of the weaker stream so that it
competes in the shared selection. Consistent with this account, smoothing the
\emph{latents} after alignment instead---which cannot change what the selection sees---helps
substantially less (Sec.~\ref{sec:results}).

\subsection{Partitioned dictionary}
\label{sec:partitioned}

A shared support is well motivated when the streams are symmetric views of the same
content, as in the vision--language setting SPARC was designed for. Spatial
transcriptomics is not symmetric: expression predicts morphology well
(cross-reconstruction NMSE $0.04$--$0.21$) whereas morphology predicts expression
poorly ($0.89$--$0.96$). Forcing every dimension to be shared therefore obliges
molecular programs with no morphological counterpart to compete for the same atoms as
programs that do have one.

We relax the constraint by partitioning the dictionary into a shared block and one
private block per stream. Global TopK selects $k_{\mathrm{shared}}$ atoms from the
shared block using the summed pre-activations, while each stream additionally selects
$k_{\mathrm{priv}}$ atoms from its own private block:
\begin{equation}
\mathcal{I}^{s}_i = \operatorname{TopK}\Bigl(\textstyle\sum_{t}\mathbf{h}^{t}_i\big|_{\mathcal{S}},\, k_{\mathrm{shared}}\Bigr)
\;\cup\;
\operatorname{TopK}\bigl(\mathbf{h}^{s}_i\big|_{\mathcal{P}_s},\, k_{\mathrm{priv}}\bigr).
\end{equation}
The private budget is carved out of $k$ rather than added to it, so each stream remains
exactly $k$-sparse and the comparison is capacity-matched. We use an equal split of
both the dictionary and the sparsity budget. Concepts that are genuinely bimodal remain
shared and interpretable across streams; modality-specific structure is no longer
forced into that shared vocabulary.

\subsection{Phase 2 and evaluation}

The concept codes are passed to MP-MNCA's masked neighbor cross-attention with the
neighbor-contrastive objective, trained for 20 epochs with pseudo-labels from
$k$-means on histology---never from anatomical annotations. Evaluation follows the
main paper exactly: PCA to 128 dimensions, a tied-covariance Gaussian mixture with the
component count set by the observed number of layers, 15-nearest-neighbor spatial
refinement, and permutation-invariant ARI/NMI. Phase 1 uses $L=1024$ dictionary atoms
with $k=128$ active, 50 epochs, batch size 256, Adam at $10^{-4}$.

\begin{figure*}[t]
\centering
\includegraphics[width=\textwidth]{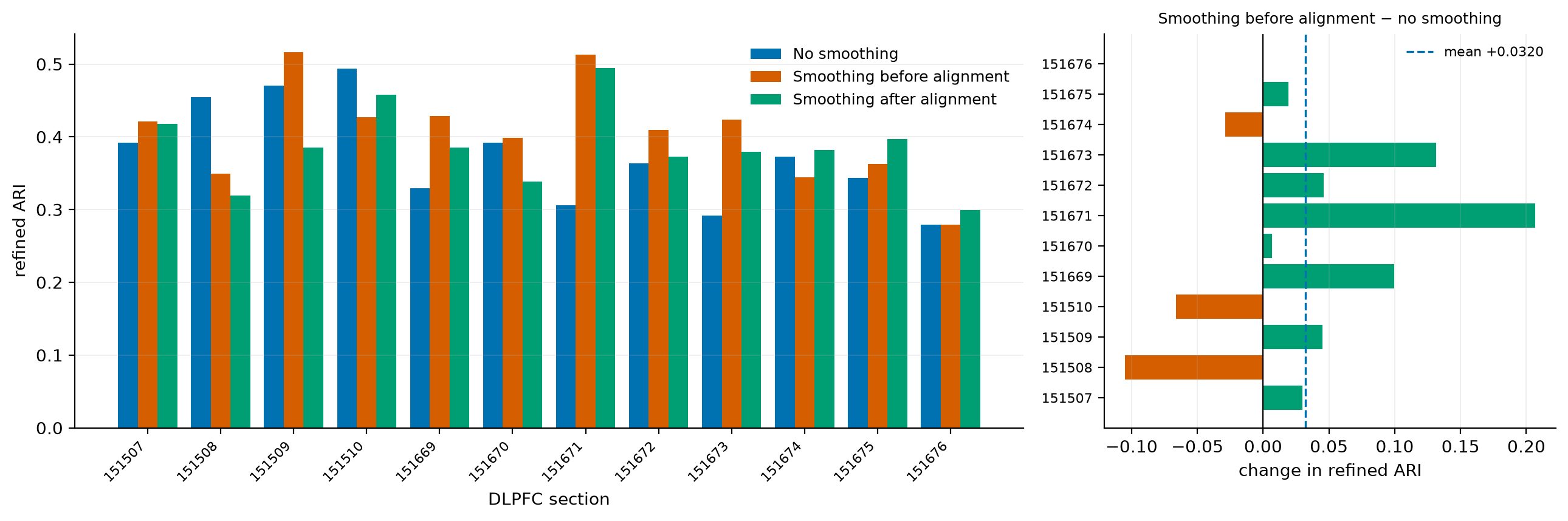}
\caption{Effect of spatial smoothing across the twelve DLPFC sections.
\textbf{Left:} refined ARI per section for no smoothing, smoothing applied before
alignment, and smoothing applied to the latents after alignment.
\textbf{Right:} per-section change from smoothing before alignment. The gain is
concentrated in sections whose sparse codes were least stable (151671, $+0.207$;
151673, $+0.132$), and four sections regress.}
\label{fig:arms}
\end{figure*}

\section{Results}
\label{sec:results}

\subsection{Spatial smoothing before alignment}

Table~\ref{tab:persection} reports per-section ARI and NMI for both support
mechanisms. Fig.~\ref{fig:arms} compares the three placements of the smoothing
operator, holding the shared support fixed. Smoothing \emph{before} alignment reaches $0.4061 \pm 0.0676$ mean refined
ARI, against $0.3740$ without smoothing and $0.3856$ when smoothing is applied to the
latents afterwards---supporting the account in Sec.~\ref{sec:method} that the
intervention acts by changing what the shared selection sees.

We report this as a promising rather than an established effect. The $+0.0320$
improvement wins $8$ of $12$ sections but does not reach significance when paired
across sections (Wilcoxon $p = 0.233$; paired $t$-test $p = 0.218$), and four sections
regress, the largest by $-0.105$. Donor structure is visible in the variance: three of
the four weakest sections belong to donor 3.

We additionally evaluated per-stream quotas and reciprocal rank fusion for the shared
selection, nonlinear encoders, per-stream logit normalization, dictionary capacity,
batch size, masked denoising, and a single dictionary pooled across all sections. All
were neutral or harmful. Only spatial smoothing and the partitioned dictionary
(Sec.~\ref{sec:partitioned}) improved clustering.

\subsection{Effect of relaxing the shared support}

Table~\ref{tab:persection} compares the two support mechanisms. The partitioned
dictionary attains $0.4228 \pm 0.0533$ mean refined ARI and $0.5285 \pm 0.0496$ NMI,
against $0.4061 \pm 0.0676$ and $0.5091 \pm 0.0501$ under a fully shared support. The
mean difference of $+0.0167$ is not significant across the twelve sections
(Wilcoxon $p = 0.519$), and the variant wins six of twelve.

Two aspects are more informative than the mean. First, the between-section standard
deviation falls to $0.0533$, the lowest of any configuration we evaluated, and the
gains are concentrated in the sections that previously performed worst---151676 moves
from $0.2789$ to $0.4074$ and 151674 from $0.3441$ to $0.4565$. This is the behaviour
the asymmetry argument predicts: sections whose molecular structure is least reflected
in morphology benefit most from private capacity. Second, NMI improves on ten of
twelve sections, a more consistent pattern than ARI alone suggests.

\begin{figure*}[t]
\centering
\includegraphics[width=\textwidth]{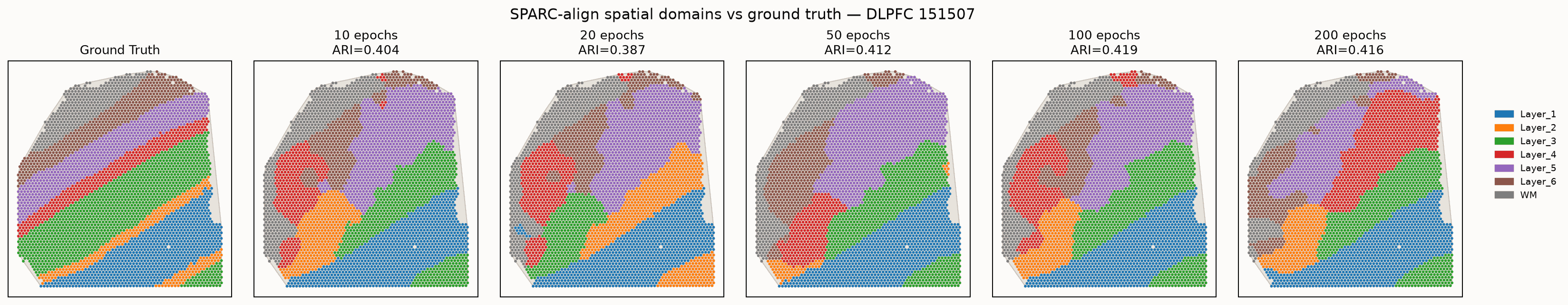}
\caption{Predicted spatial domains on section 151507 under the partitioned dictionary,
at 10, 20, 50, 100 and 200 Phase-1 epochs, with ground truth at left. Laminar
organization is established by the earliest checkpoint and remains stable thereafter.}
\label{fig:annot_partitioned}
\end{figure*}

Fig.~\ref{fig:annot_partitioned} shows the corresponding convergence on section
151507, which is stable across epoch budgets ($0.404$ at 10 epochs to $0.416$ at 200).

\subsection{Convergence and stability}

\begin{figure*}[t]
\centering
\includegraphics[width=0.92\textwidth]{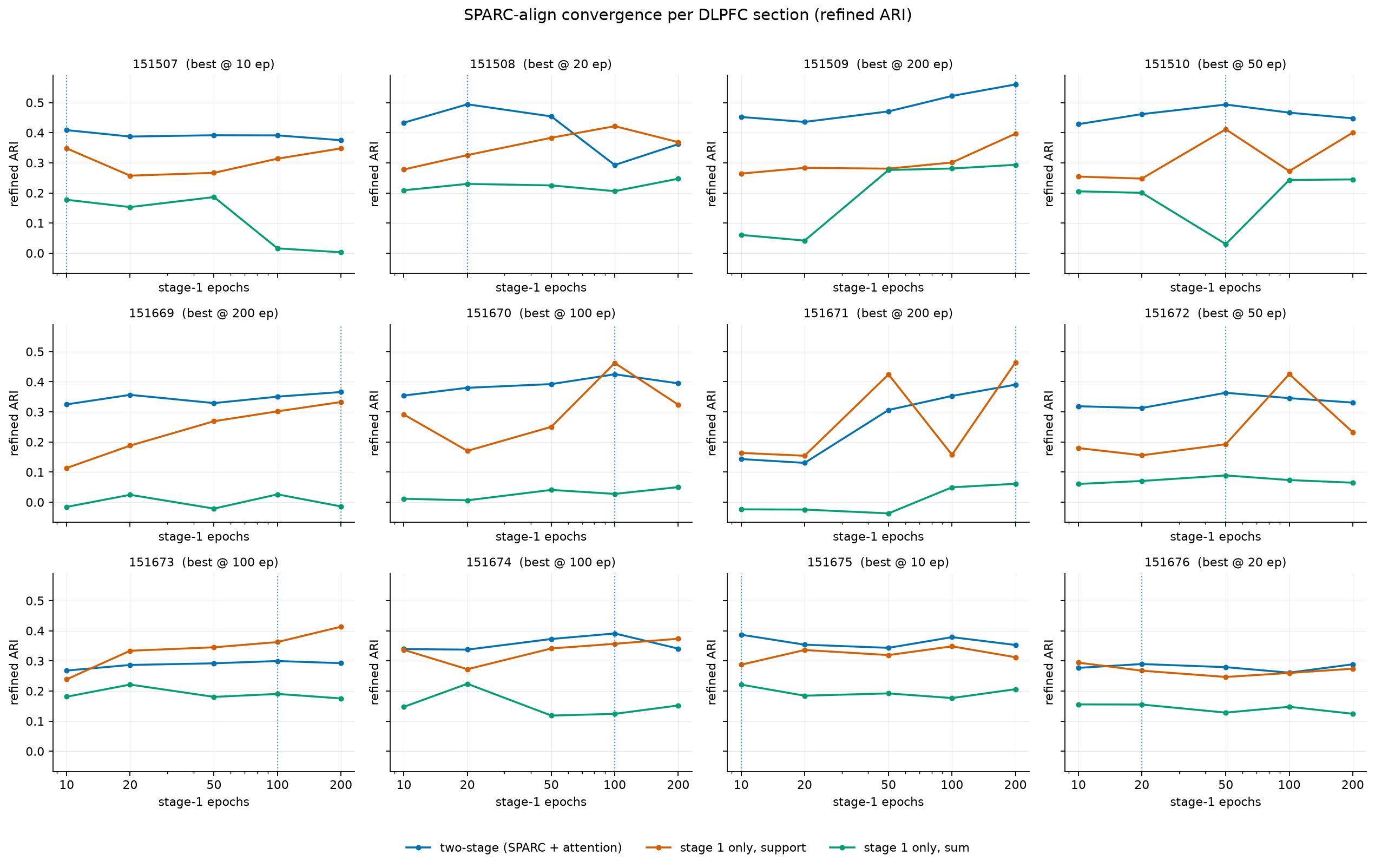}
\caption{Refined ARI against Phase-1 epochs for each section, with Phase 2 fixed at 20
epochs. Blue is the full two-phase pipeline; orange and green are Phase 1 alone under
two cluster-input choices. The dotted line marks each section's best epoch. All
checkpoints come from a single 200-epoch run per section, so the curves are
trajectories rather than independent runs.}
\label{fig:convergence}
\end{figure*}

\begin{figure}[t]
\centering
\includegraphics[width=0.48\columnwidth]{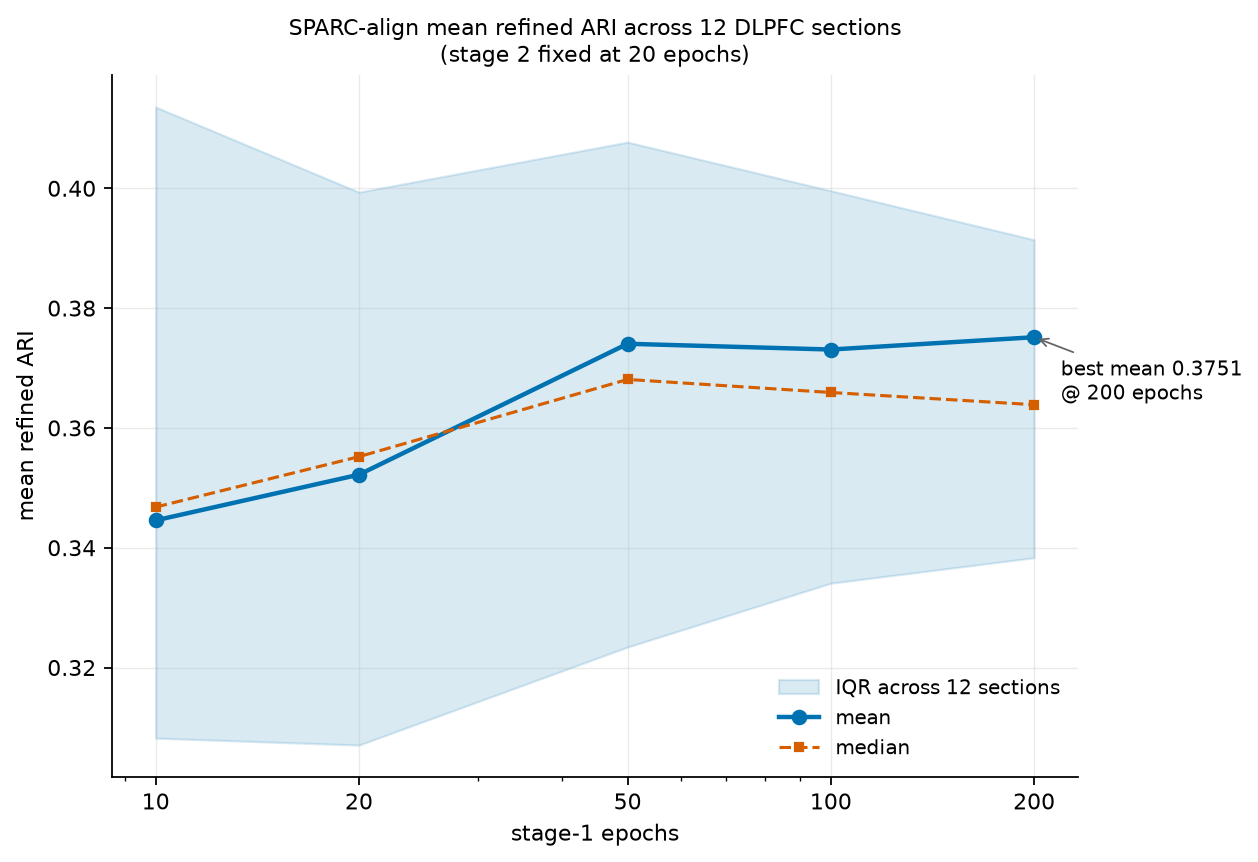}
\hfill
\includegraphics[width=0.48\columnwidth]{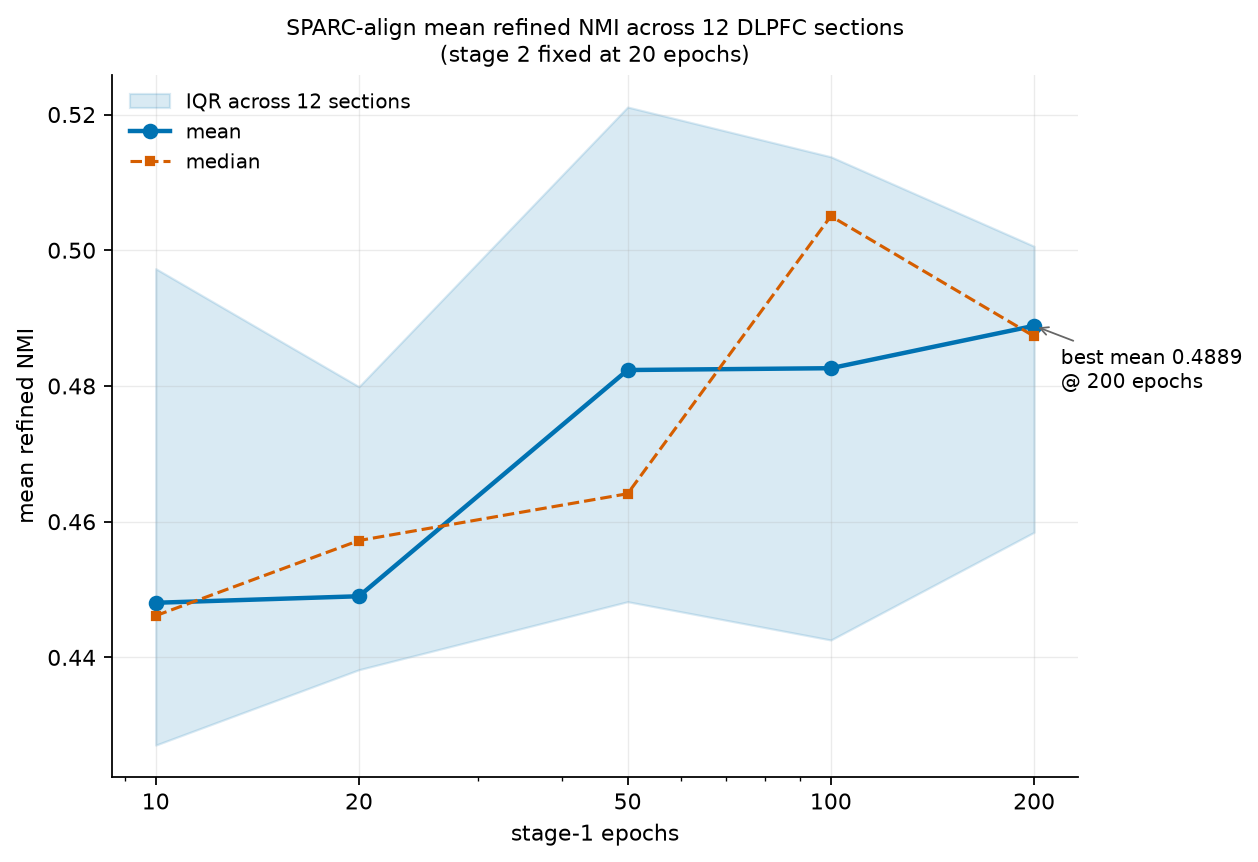}
\caption{Mean refined ARI (left) and NMI (right) across the twelve sections against
Phase-1 epochs, with an interquartile band. The mean changes by $0.031$ ARI over a
twenty-fold range of training.}
\label{fig:convergence_mean}
\end{figure}

Fig.~\ref{fig:convergence} and Fig.~\ref{fig:convergence_mean} show refined ARI as a
function of Phase-1 epochs. Two observations stand out. First, the mean varies by only
$0.031$ across a twenty-fold range of training, and every epoch budget tested is
optimal for at least one section, so no transferable epoch setting exists. Second, the
two-phase pipeline is markedly more stable than Phase 1 alone: across all
(section, epoch) checkpoints it attains higher mean ARI ($0.364$ vs.\ $0.300$) at half
the within-section volatility ($0.031$ vs.\ $0.063$). Phase 2 therefore contributes
stabilization as much as accuracy, smoothing a sparse code that is otherwise erratic
across checkpoints.

\begin{figure}[t]
\centering
\includegraphics[width=\columnwidth]{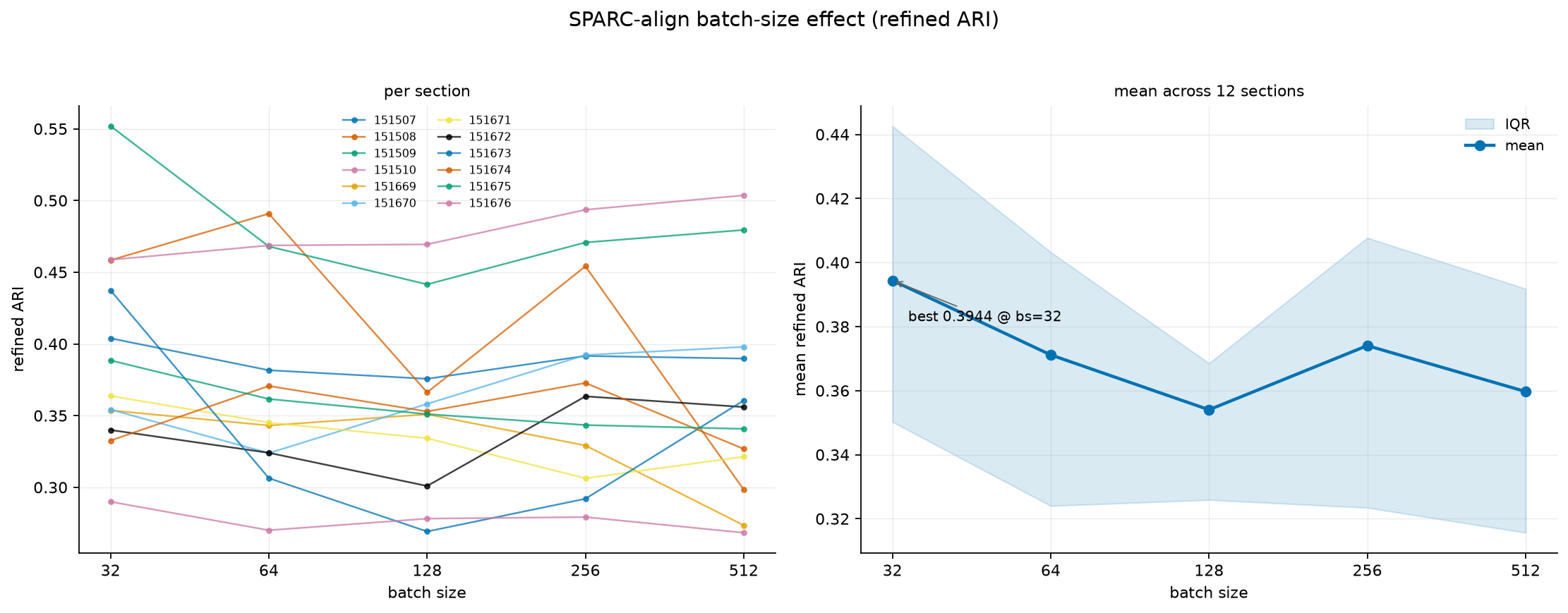}
\caption{Batch-size sweep at a fixed epoch budget. Left: per section. Right: mean with
interquartile band. The spread over a sixteen-fold range is $0.040$ ARI and the
nominal optimum is not significant.}
\label{fig:batch}
\end{figure}

Fig.~\ref{fig:batch} shows the batch-size sweep. Across a sixteen-fold range the total
spread is $0.040$ ARI, and the nominal optimum is not significant
(Wilcoxon $p = 0.30$). Which section is being processed matters roughly twice as much
as which batch size is chosen.

\subsection{Qualitative behavior}

\begin{figure*}[t]
\centering
\includegraphics[width=\textwidth]{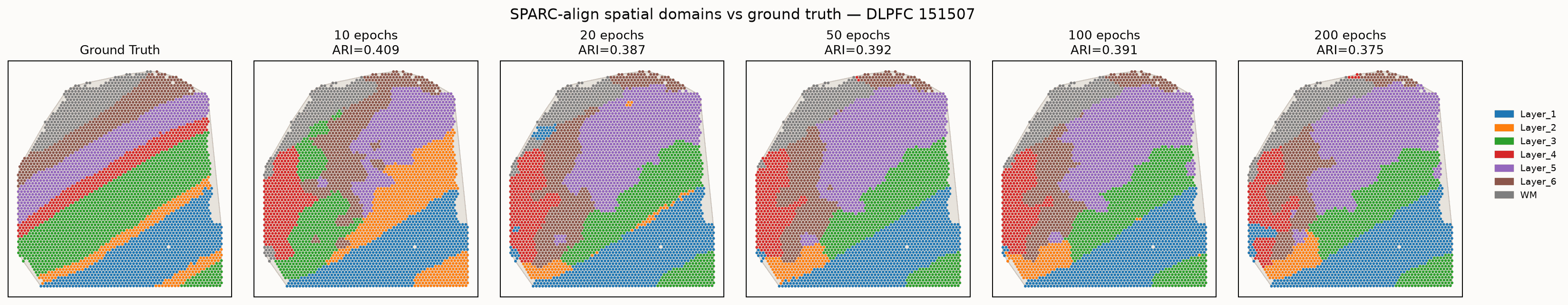}\\[2pt]
\includegraphics[width=\textwidth]{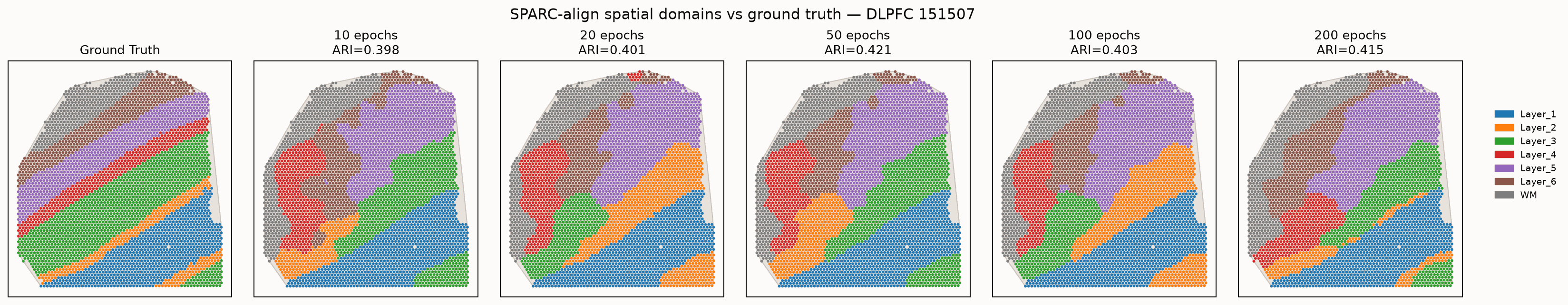}
\caption{Predicted spatial domains on DLPFC section 151507 at 10, 20, 50, 100 and 200
Phase-1 epochs, without smoothing (\textbf{top}) and with smoothing before alignment
(\textbf{bottom}); ground truth at left. Smoothing converts spatially incoherent
predictions into laminar bands, but Layer~4 remains too thick and white matter stays
entangled with Layer~6 at every budget.}
\label{fig:annot}
\end{figure*}

Fig.~\ref{fig:annot} shows predicted domains on section 151507 across training, with
and without smoothing. Smoothing changes the character of the failure. Without it,
predictions are spatially incoherent blobs; with it, the model recovers recognizable
laminar bands that follow the ground-truth orientation and sharpen with training.
The residual errors are systematic rather than random: Layer~4---the thinnest cortical
layer---is consistently rendered too thick, and white matter remains entangled with
Layer~6. Smoothing therefore corrects the \emph{geometry} of the prediction without
resolving layer \emph{identity} at boundaries.

\begin{table}[t]
\centering
\small
\setlength{\tabcolsep}{5pt}
\begin{tabular}{ll cc cc}
\toprule
& & \multicolumn{2}{c}{Shared support} & \multicolumn{2}{c}{Partitioned} \\
\cmidrule(lr){3-4}\cmidrule(lr){5-6}
Section & Donor & ARI & NMI & ARI & NMI \\
\midrule
151507 & 1 & 0.4214 & 0.5216 & 0.4125 & 0.5344 \\
151508 & 1 & 0.3491 & 0.4635 & \textbf{0.4451} & 0.5450 \\
151509 & 1 & \textbf{0.5159} & 0.5710 & 0.5082 & 0.5816 \\
151510 & 1 & 0.4272 & 0.5267 & 0.4558 & 0.5095 \\
151669 & 2 & 0.4287 & 0.5202 & 0.3678 & 0.4968 \\
151670 & 2 & 0.3990 & 0.4546 & 0.3124 & 0.3965 \\
151671 & 2 & 0.5131 & \textbf{0.5942} & 0.4772 & 0.5869 \\
151672 & 2 & 0.4093 & 0.5112 & 0.4397 & 0.5588 \\
151673 & 3 & 0.4235 & 0.5549 & 0.4162 & 0.5469 \\
151674 & 3 & 0.3441 & 0.4616 & \textbf{0.4565} & 0.5371 \\
151675 & 3 & 0.3626 & 0.5010 & 0.3745 & 0.5107 \\
151676 & 3 & 0.2789 & 0.4289 & \textbf{0.4074} & 0.5383 \\
\midrule
\multicolumn{2}{l}{Mean $\pm$ std} & $0.4061$ & $0.5091$ & $\mathbf{0.4228}$ & $\mathbf{0.5285}$ \\
\multicolumn{2}{l}{}               & $\pm 0.0676$ & $\pm 0.0501$ & $\pm \mathbf{0.0533}$ & $\pm 0.0496$ \\
\multicolumn{2}{l}{Median}         & 0.4154 & 0.5157 & \textbf{0.4279} & \textbf{0.5377} \\
\bottomrule
\end{tabular}
\caption{SPARC-align with spatial smoothing on the twelve DLPFC sections, under a
shared concept support (Sec.~\ref{sec:method}) and under the partitioned dictionary
(Sec.~\ref{sec:partitioned}). Standard deviation uses $\mathrm{ddof}=1$, treating
sections as the biological replicates. Bold marks the larger of the two where the
difference exceeds $0.05$.}
\label{tab:persection}
\end{table}

\section{Discussion, limitations and future work}
\label{sec:discussion}

\paragraph{A modality gap bounds the approach.}
SPARC-align improves substantially on the published SPARC configuration
($0.2662 \rightarrow 0.4228$ mean refined ARI) but does not reach MP-MNCA. The
limitation appears structural rather than one of tuning. Cross-reconstruction from
histology to expression never falls below $\approx 0.89$ NMSE in any configuration we
tested, meaning morphology cannot predict expression in this tissue; the shared
concept space consequently remains largely a histological code, while laminar identity
resides in the molecular channel. A concept-agreement diagnostic makes the same point:
neighboring spots share four to five times more active concepts than random pairs
(Jaccard $0.29$ vs.\ $0.06$), yet same-layer and cross-layer neighbors differ by only
$\approx 0.02$ (AUC $\approx 0.56$). The learned concepts capture spatial structure
without capturing layer boundaries---precisely the behavior visible in
Fig.~\ref{fig:annot}.

\paragraph{Reconstruction is not the binding constraint.}
It is natural to attribute the gap to poor reconstruction of the expression stream. We
find this is not the case. Across the epoch range, gene self-NMSE improves by $15.4\%$
while ARI improves by $8.9\%$; pooling a single dictionary over all twelve sections
raises data per atom elevenfold yet moves self-NMSE only from $0.880$ to $0.872$; and
enlarging the encoder degrades reconstruction under a fixed budget, indicating
undertraining rather than insufficient capacity. Improving reconstruction fidelity is
therefore an inefficient route to better spatial domains.

\paragraph{Relaxing the shared support helps, and points forward.}
Of every modification we evaluated, only two improved clustering: spatial smoothing
and the partitioned dictionary. Both act on the same underlying issue---that a
histology-dominated shared code carries limited laminar information---and the
partitioned variant additionally reduces between-section variance while raising NMI on
ten of twelve sections. Neither reaches significance individually at $n = 12$, but they
are consistent in direction and mechanism, which suggests that the productive axis is
how much representational independence the two modalities are granted, rather than how
the shared selection is weighted. A control that weighted smoothing by expression
similarity instead of histology was significantly \emph{worse}
(Wilcoxon $p = 0.021$), indicating that the benefit of the morphology prior derives
from its independence from the stream being smoothed.

\paragraph{Limitations.}
(i) Neither principal effect is statistically established at $n = 12$
($p = 0.233$ for smoothing, $p = 0.519$ for partitioning). With twelve sections and
differences of this magnitude, the achievable power is low; the pipeline is
bit-reproducible at a fixed seed, so additional seeds would reduce per-section
measurement noise but would not increase the number of biological replicates.
Confirmation requires further sections or a second tissue.
(ii) $\alpha = 0.5$ and $\beta = 1$ were set by analogy, not tuned. (iii) Smoothing
alters the reconstruction target, so NMSE is not comparable between smoothed and
unsmoothed variants; only clustering metrics are. (iv) Results are per-section
transductive on a single tissue type; generalization to held-out donors or other
tissues is untested. (v) The component count is set from the observed number of
annotated layers, so model selection is label-count-informed, as in the main paper.

\paragraph{Future work.}
The most informative next step is evaluation on additional tissue, since the effects we
observe are not resolvable at twelve sections regardless of how many seeds are averaged
(Limitations, (i)). Beyond that, the diagnostics point away from the
sparse-reconstruction formulation itself. A reconstruction objective allocates capacity
to variance, and laminar identity is evidently not where the variance lies; an
objective that is discriminative in the concept space appears more promising than
further refinement of the shared-support mechanism. The partitioned dictionary suggests
a related direction: its shared and private blocks were split evenly without tuning,
and the balance between them is a natural axis to explore. Validating
dictionary atoms against published laminar marker genes would also test concept
quality directly, without relying on the annotations used for evaluation.

\section{Section-wise Convergence Analysis}
\label{sec:convergence_analysis}

\begin{figure*}[t]
    \centering
    \includegraphics[width=\textwidth]{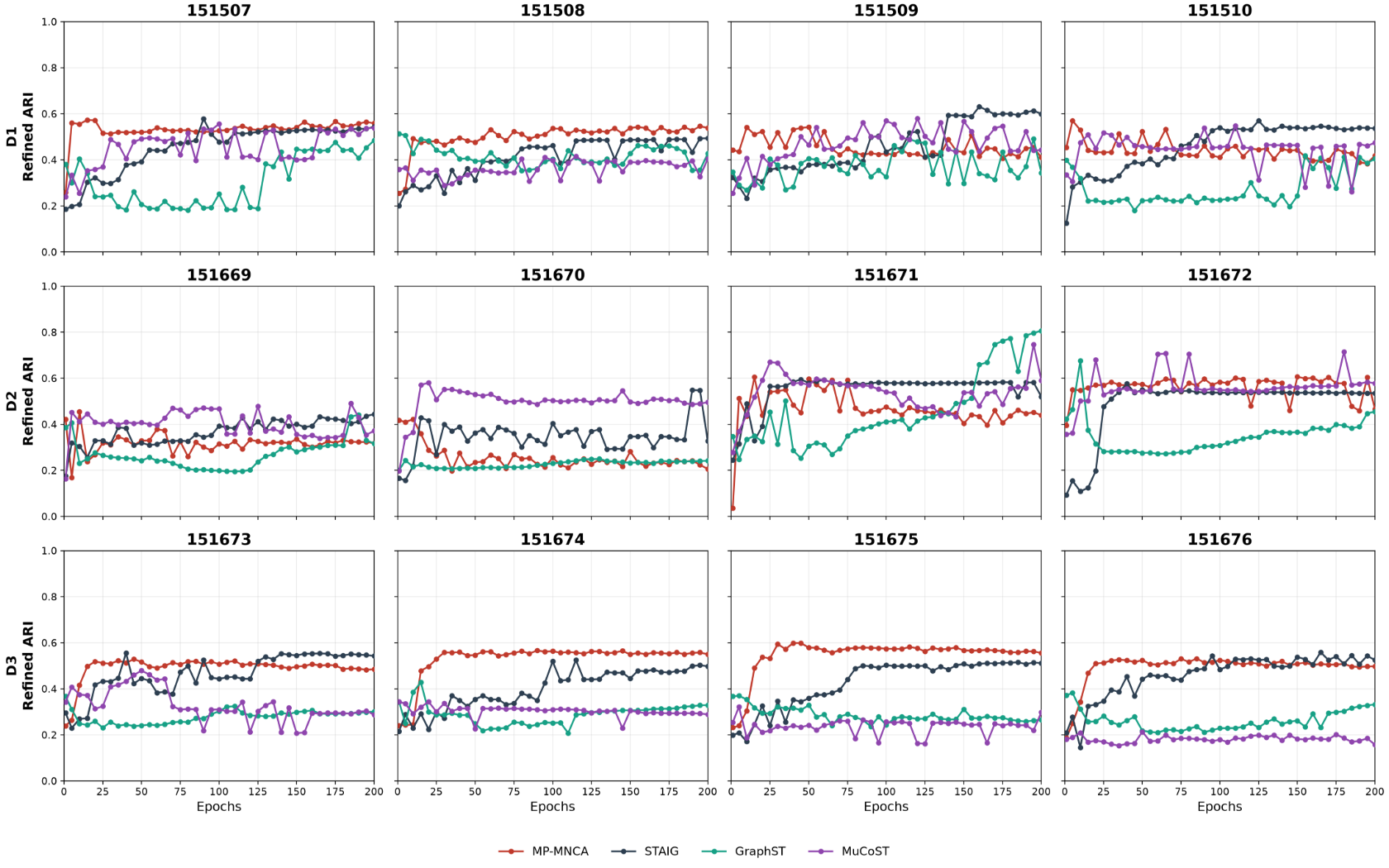}
    \caption{
    \textbf{Section-wise convergence of refined ARI on the 12 DLPFC sections.}
    Each panel reports refined ARI over the common 200-epoch optimization
    budget for MP-MNCA, STAIG, GraphST, and MuCoST. MP-MNCA typically reaches
    a competitive regime early in training, whereas the competing methods
    exhibit more section-dependent convergence behavior.
    }
    \label{fig:section_convergence}
\end{figure*}

To examine optimization behavior beyond a single final checkpoint, we track
the refined ARI of all methods throughout the common 200-epoch training
budget on each of the 12 DLPFC sections. Figure~\ref{fig:section_convergence}
shows the complete section-wise trajectories. The results reveal substantial
variation in optimization dynamics across tissue sections. MP-MNCA generally
enters a competitive high-ARI regime within the early stages of training and
then remains comparatively stable. In contrast, STAIG often improves more
gradually and reaches competitive performance later in training, while
GraphST and MuCoST exhibit larger section-dependent fluctuations. These
trajectories therefore distinguish convergence behavior that would be
obscured by reporting only a final or best checkpoint.

\begin{figure}[h]
    \centering
    \includegraphics[width=\textwidth]{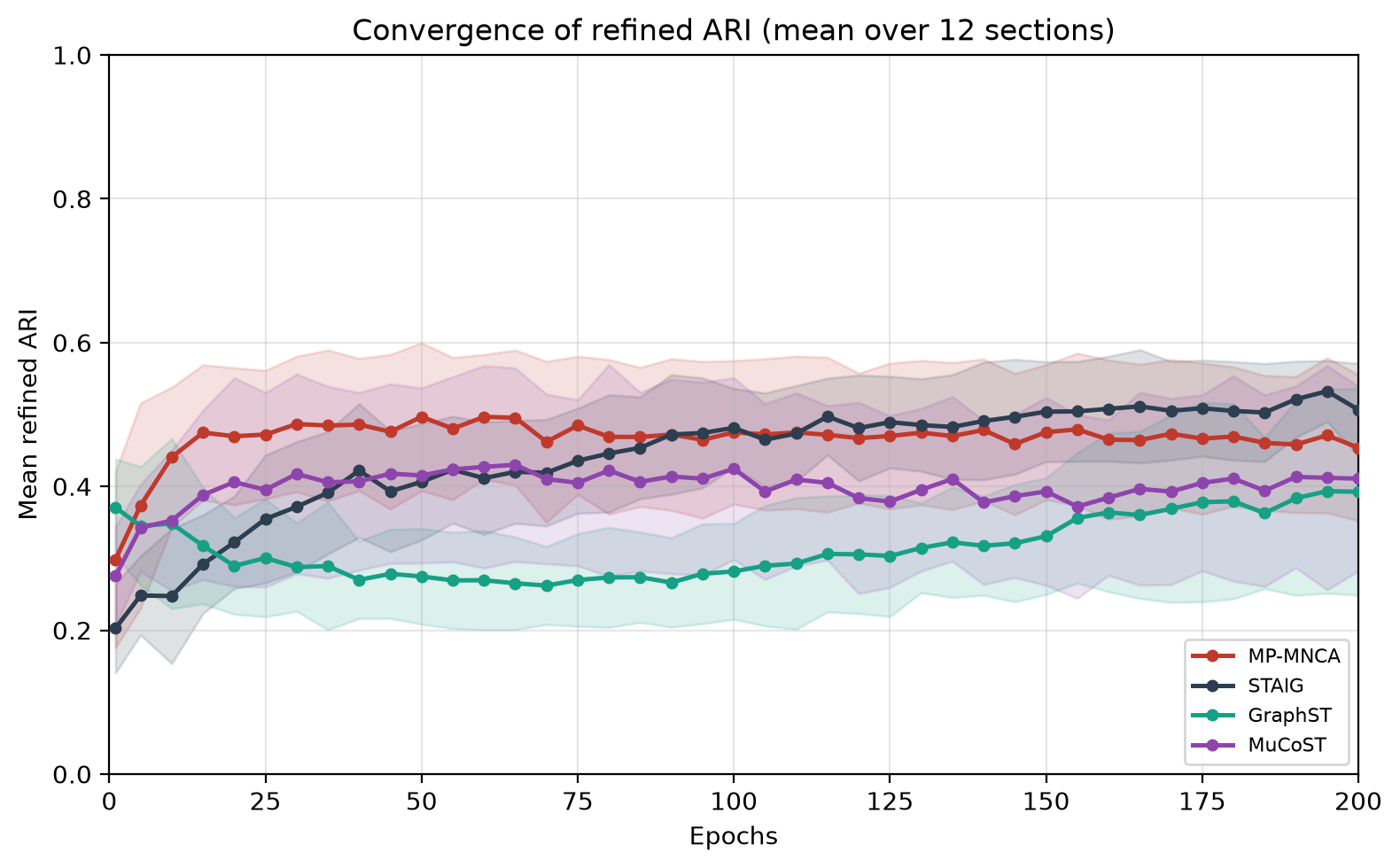}
    \caption{
    \textbf{Mean convergence of refined ARI across the 12 DLPFC sections.}
    Curves denote the mean refined ARI at each epoch and shaded regions denote
    $\pm$ one standard deviation across sections. MP-MNCA reaches a competitive
    high-ARI regime early in optimization, while STAIG improves more gradually
    and becomes comparable later in training.
    }
    \label{fig:mean_convergence}
\end{figure}

To summarize these section-level trajectories, we additionally aggregate
performance across all 12 sections at every epoch. Figure~\ref{fig:mean_convergence}
reports the mean refined ARI together with the corresponding standard
deviation across sections. The aggregate view confirms that MP-MNCA reaches
a competitive performance regime within the first few tens of epochs,
whereas STAIG improves more progressively and becomes comparable later in
training. The shaded regions further capture cross-section variability,
complementing the individual trajectories in
Figure~\ref{fig:section_convergence}. Importantly, this comparison measures
\emph{epoch-wise optimization dynamics} under an identical training budget
rather than wall-clock computational efficiency.

\section{Epoch-wise Spatial-Domain Evolution Across DLPFC Sections}
\label{sec:supp_epochwise_maps}

To complement the quantitative convergence analysis, we visualize the
evolution of predicted spatial domains throughout training on all 12 DLPFC
sections. For each section, rows correspond to checkpoints at epochs
10, 20, 50, 100, and 200, while columns show the ground-truth annotation
and predictions from STAIG, GraphST, MuCoST, and \method.
These visualizations expose differences in optimization dynamics that are
not captured by a single final-checkpoint metric, including the emergence
of anatomically coherent laminar structure, boundary stabilization,
fragmentation, and local spatial inconsistencies.

\begin{figure*}[h]
    \centering
    \includegraphics[width=\textwidth]
    {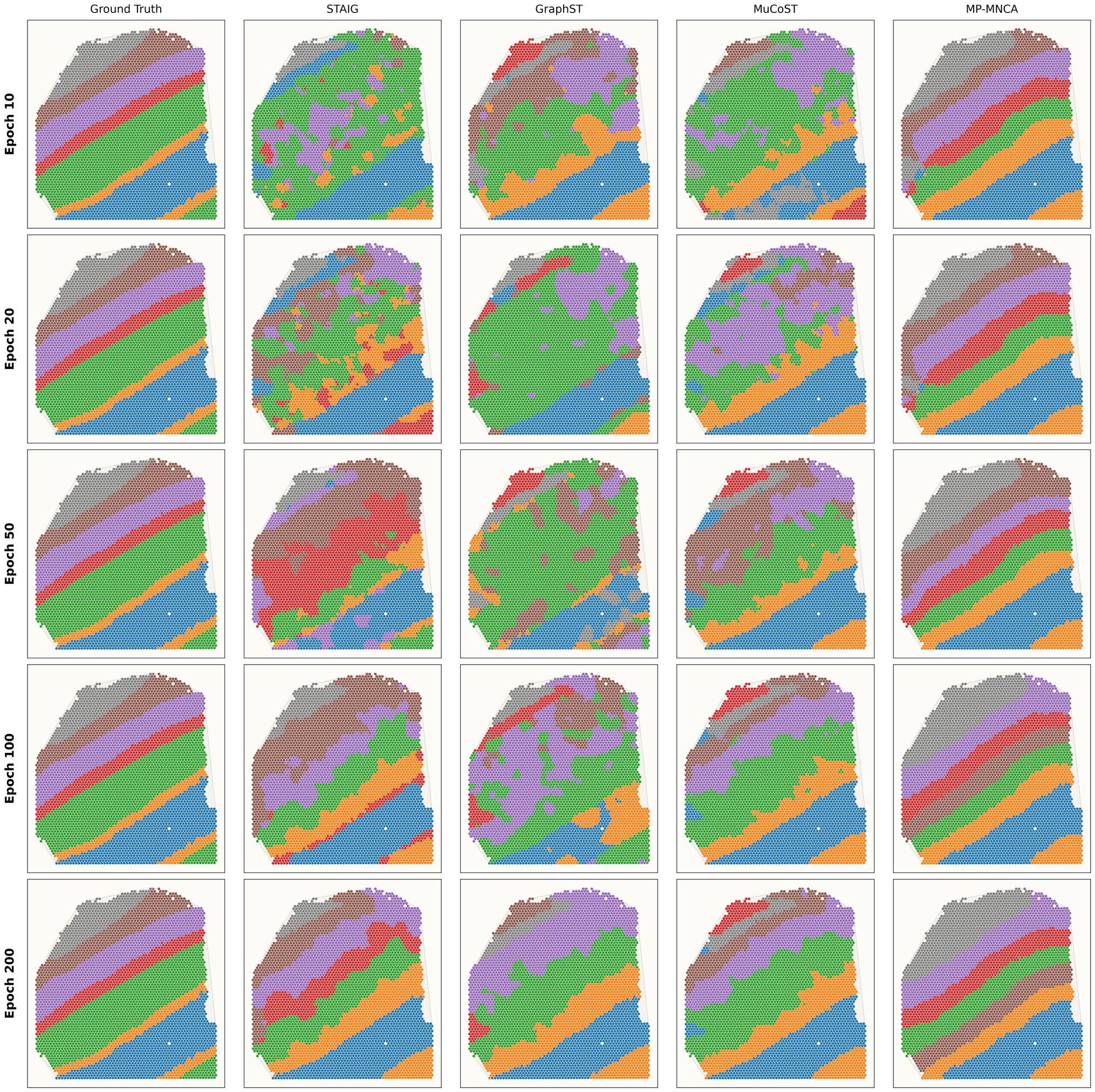}
    \caption{
    \textbf{Epoch-wise spatial-domain evolution on DLPFC section 151507.}
    Predictions are shown at epochs 10, 20, 50, 100, and 200 for STAIG,
    GraphST, MuCoST, and \method, together with the ground-truth cortical
    layers. The sequence highlights how the spatial organization and
    cortical boundaries produced by each method evolve throughout
    optimization.
    }
    \label{fig:epoch_151507}
\end{figure*}

\newpage
\begin{figure*}[h]
    \centering
    \includegraphics[width=\textwidth]
    {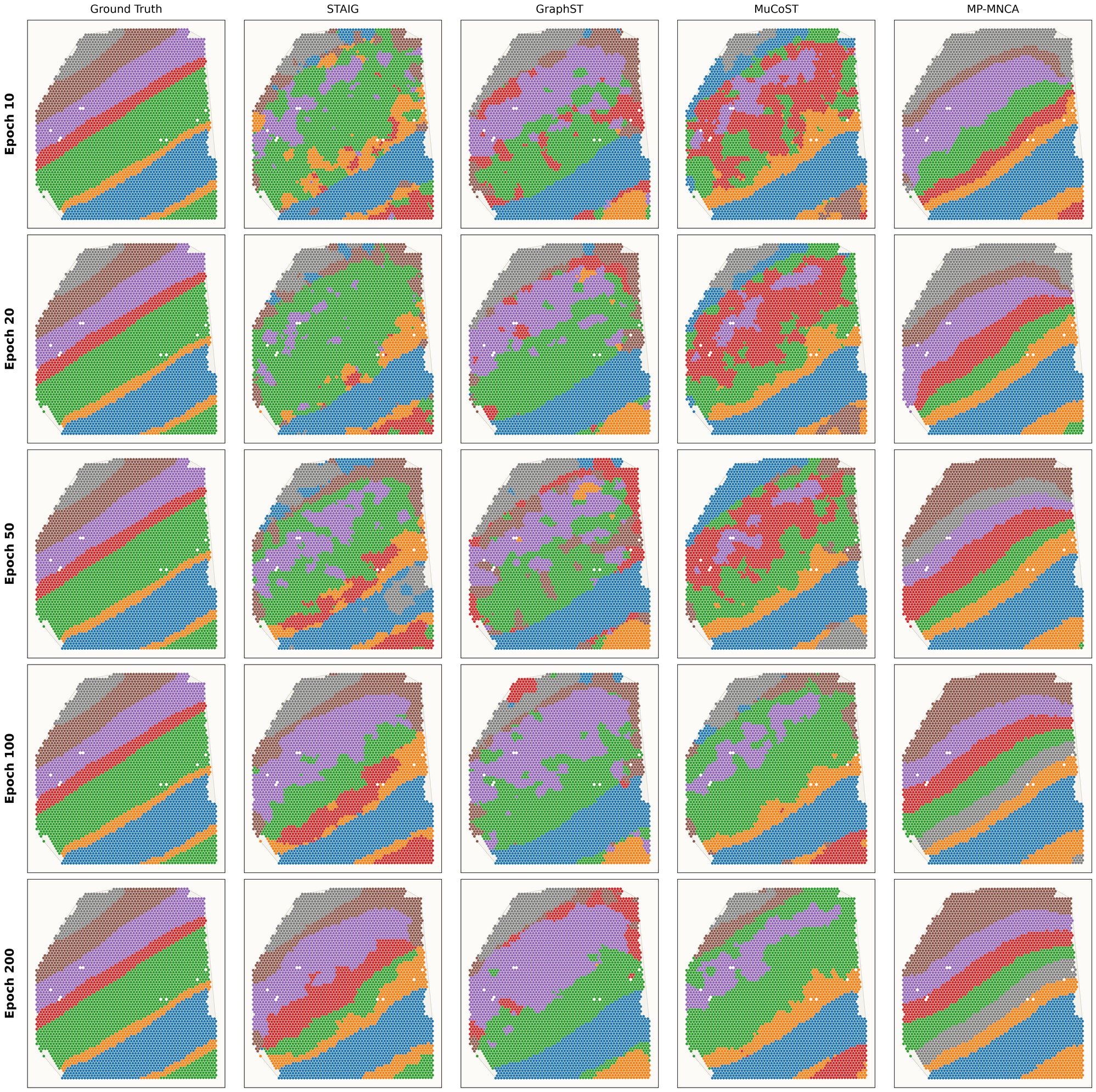}
    \caption{
    \textbf{Epoch-wise spatial-domain evolution on DLPFC section 151508.}
    Spatial-domain predictions at epochs 10, 20, 50, 100, and 200 illustrate
    the progression from early optimization to the later training regime
    for STAIG, GraphST, MuCoST, and \method.
    }
    \label{fig:epoch_151508}
\end{figure*}

\newpage
\begin{figure*}[h]
    \centering
    \includegraphics[width=\textwidth]
    {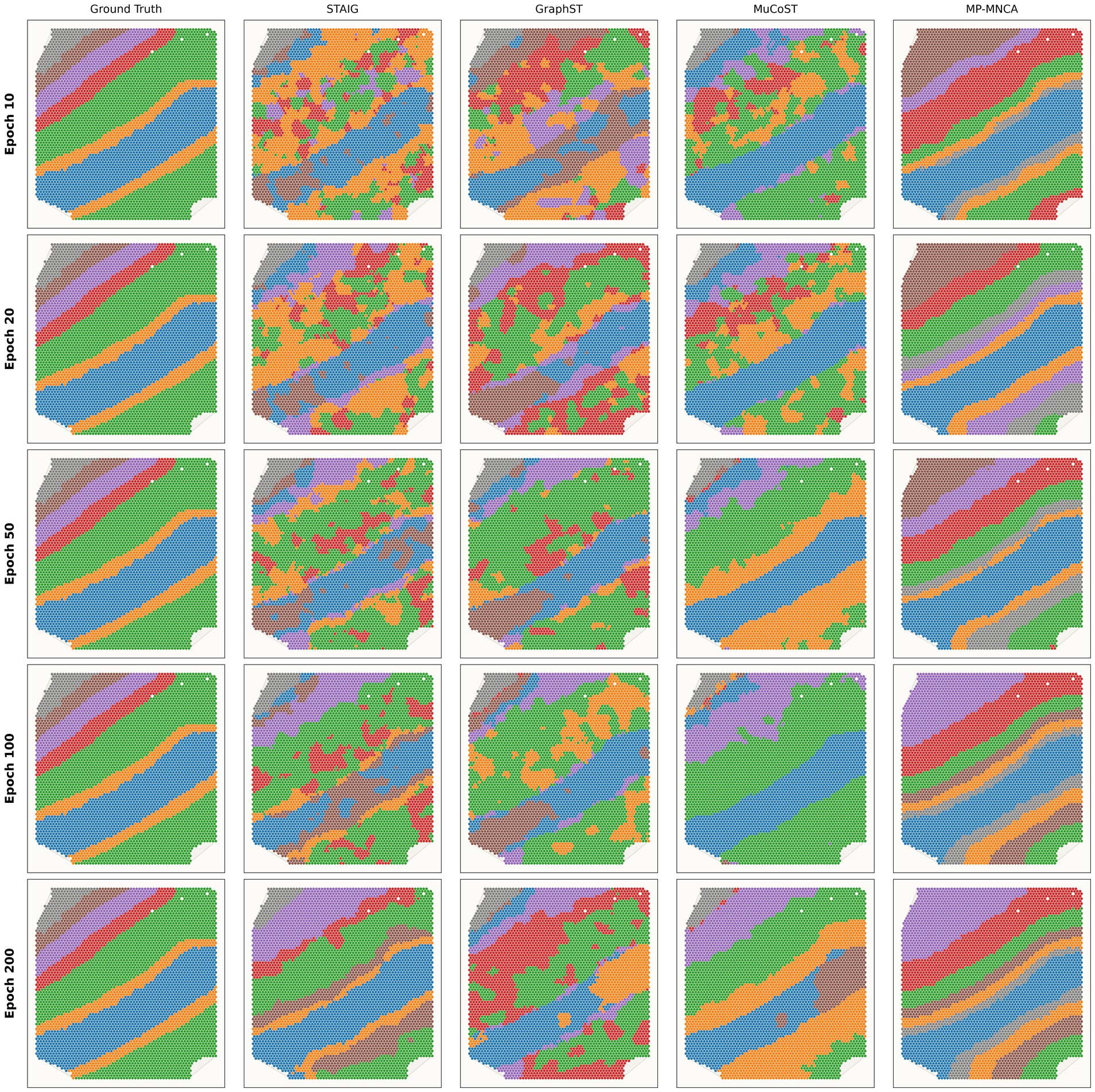}
    \caption{
    \textbf{Epoch-wise spatial-domain evolution on DLPFC section 151509.}
    This comparatively challenging section illustrates that convergence
    behavior is not uniform across tissue sections. The maps show the
    evolution of layer continuity, boundary placement, and local
    fragmentation for each method over the common 200-epoch budget.
    }
    \label{fig:epoch_151509}
\end{figure*}

\newpage
\begin{figure*}[h]
    \centering
    \includegraphics[width=\textwidth]
    {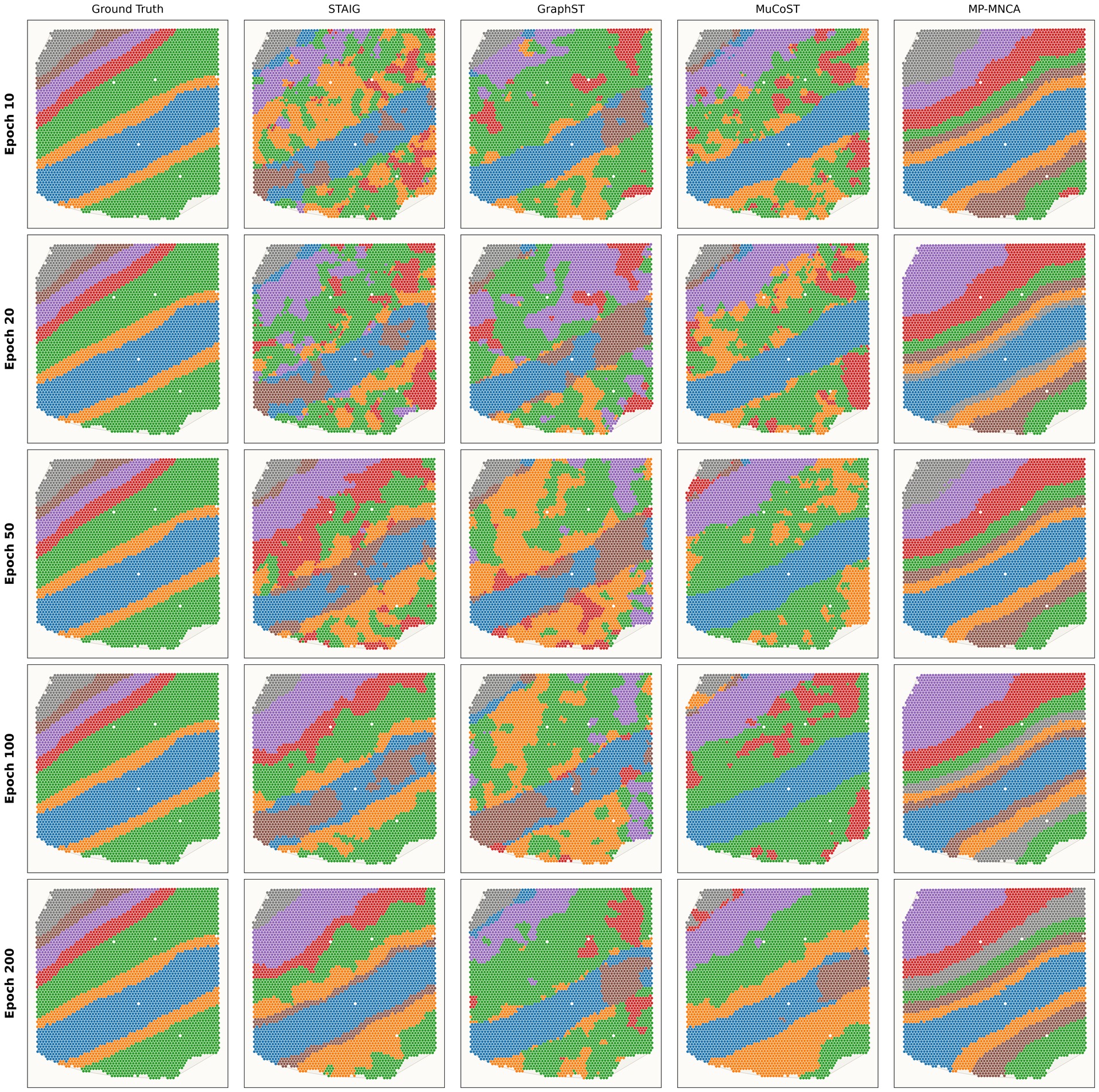}
    \caption{
    \textbf{Epoch-wise spatial-domain evolution on DLPFC section 151510.}
    Predictions across the five training checkpoints provide a qualitative
    view of how cortical-layer organization develops and stabilizes under
    STAIG, GraphST, MuCoST, and \method.
    }
    \label{fig:epoch_151510}
\end{figure*}

\newpage
\begin{figure*}[h]
    \centering
    \includegraphics[width=\textwidth]
    {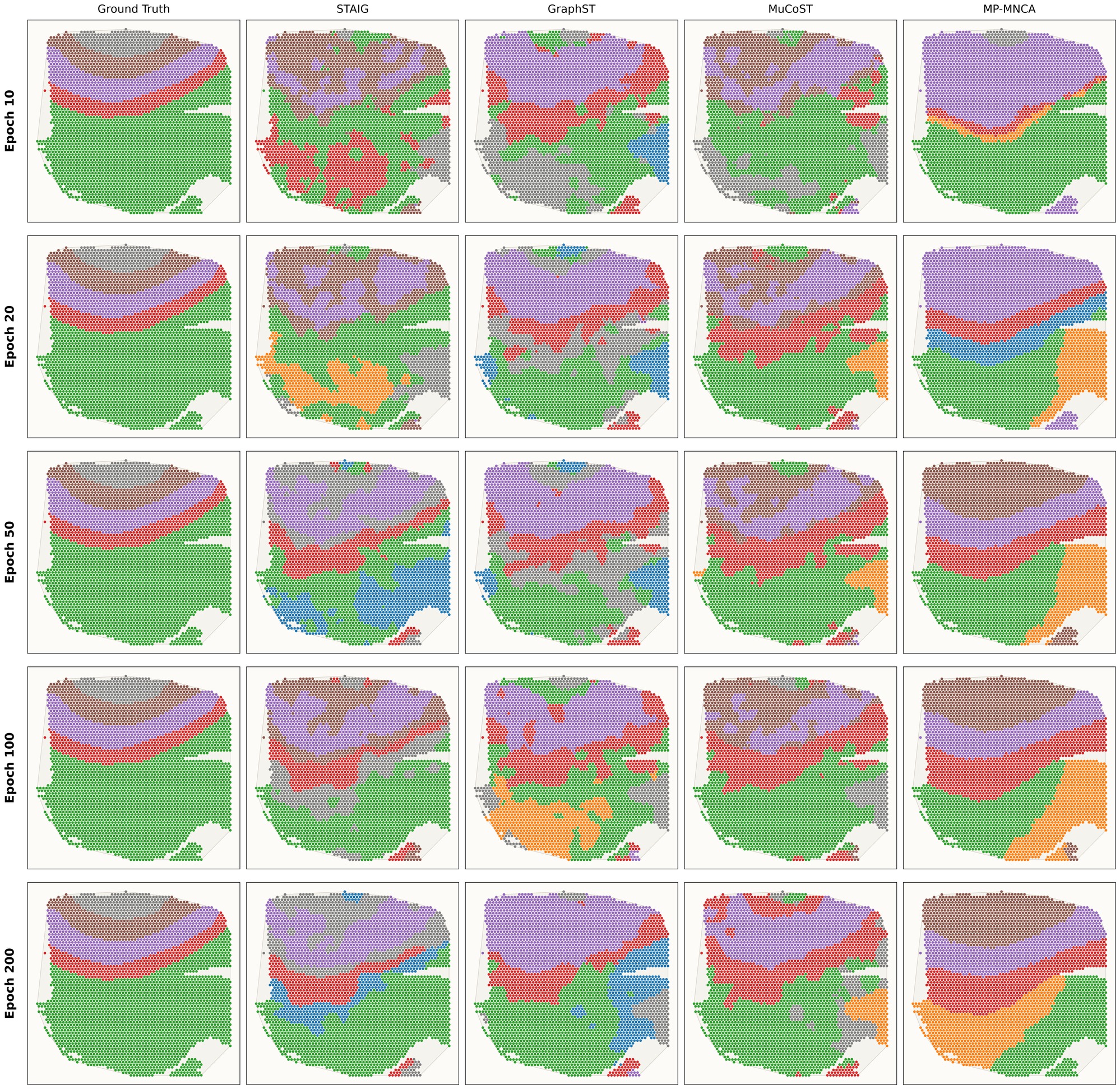}
    \caption{
    \textbf{Epoch-wise spatial-domain evolution on DLPFC section 151669.}
    The checkpoint sequence compares the formation and stabilization of
    spatial domains across competing methods, with the ground-truth
    laminar organization shown for reference.
    }
    \label{fig:epoch_151669}
\end{figure*}

\newpage
\begin{figure*}[h]
    \centering
    \includegraphics[width=\textwidth]
    {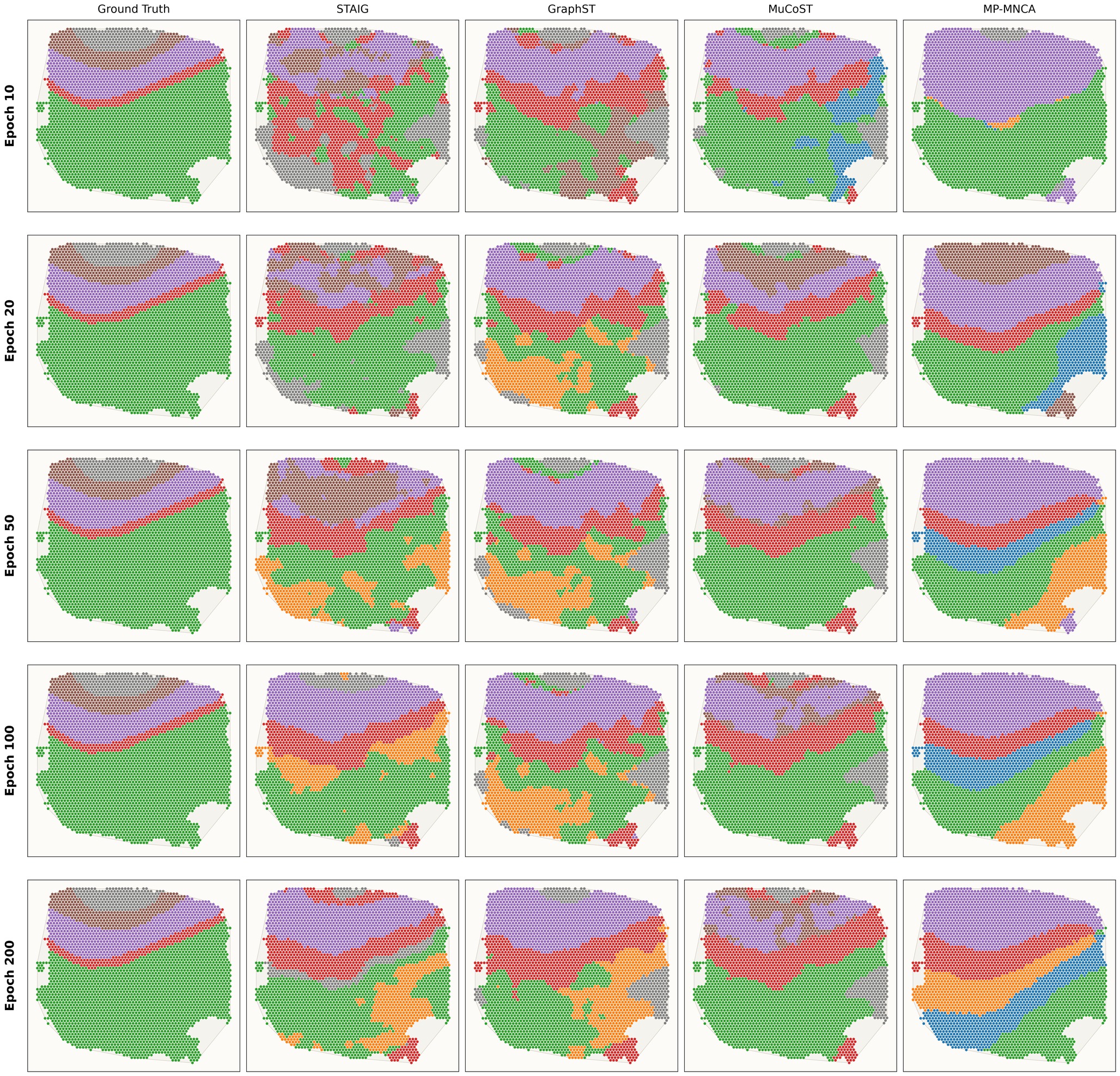}
    \caption{
    \textbf{Epoch-wise spatial-domain evolution on DLPFC section 151670.}
    Spatial-domain maps at epochs 10--200 reveal the temporal evolution of
    cortical boundaries and within-layer consistency for STAIG, GraphST,
    MuCoST, and \method.
    }
    \label{fig:epoch_151670}
\end{figure*}

\begin{figure*}[h]
    \centering
    \includegraphics[width=\textwidth]
    {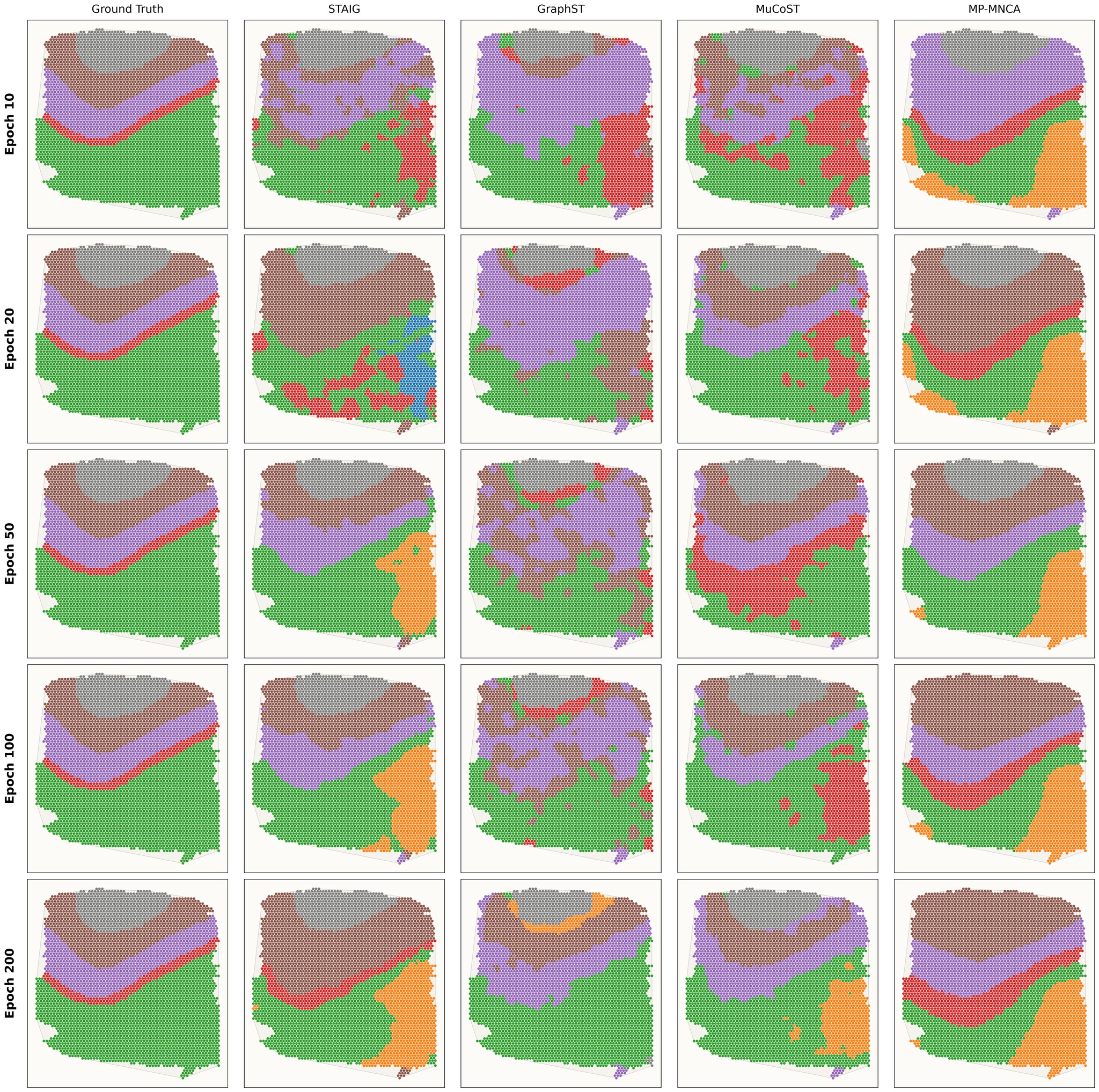}
    \caption{
    \textbf{Epoch-wise spatial-domain evolution on DLPFC section 151671.}
    The visualization tracks changes in predicted cortical organization
    throughout the shared training budget and complements the corresponding
    quantitative convergence measurements.
    }
    \label{fig:epoch_151671}
\end{figure*}

\begin{figure*}[h]
    \centering
    \includegraphics[width=\textwidth]
    {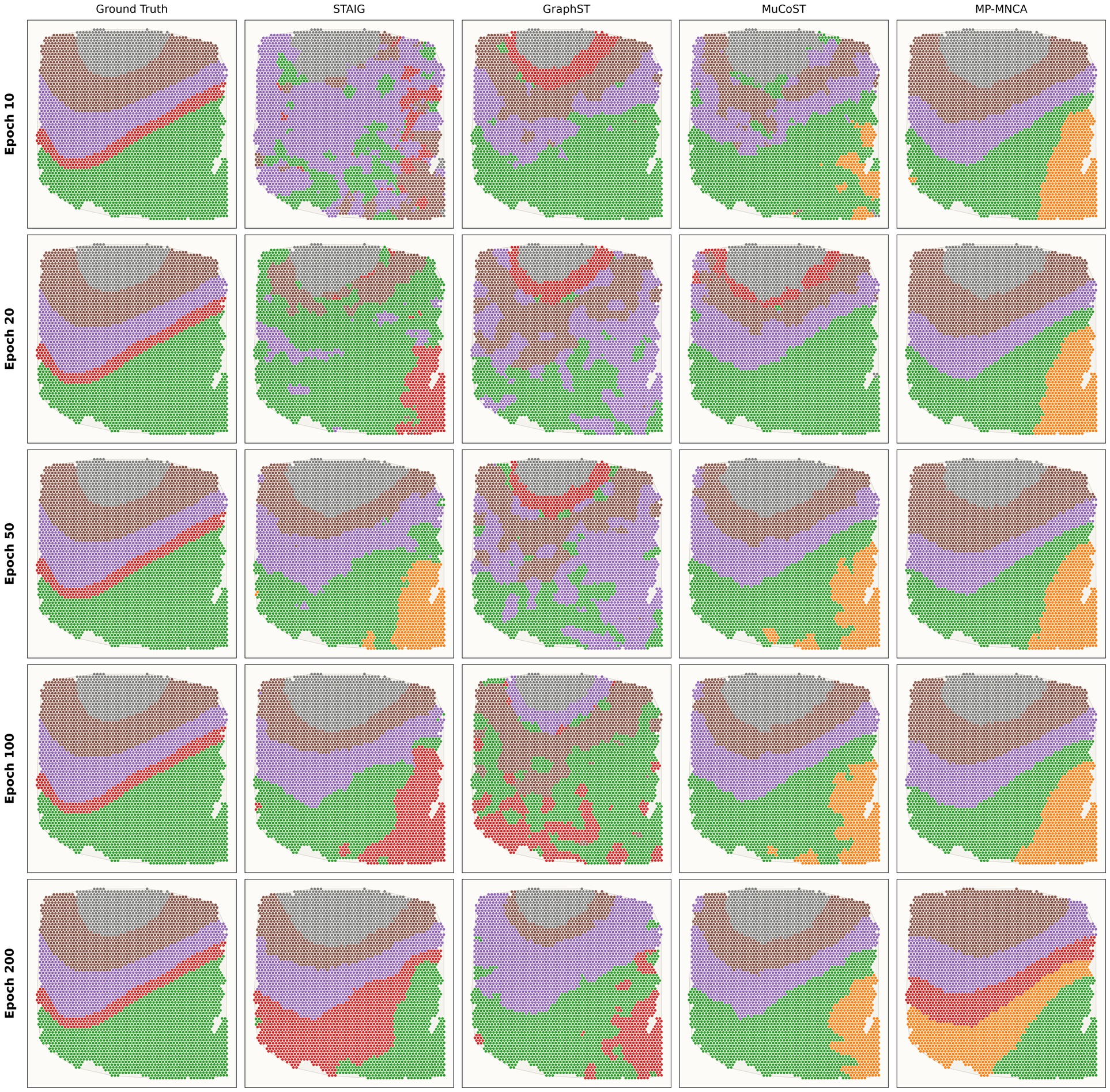}
    \caption{
    \textbf{Epoch-wise spatial-domain evolution on DLPFC section 151672.}
    Predictions from successive checkpoints illustrate differences between
    methods in the development of coherent laminar structure and the
    stabilization of spatial-domain boundaries.
    }
    \label{fig:epoch_151672}
\end{figure*}

\begin{figure*}[h]
    \centering
    \includegraphics[width=\textwidth]
    {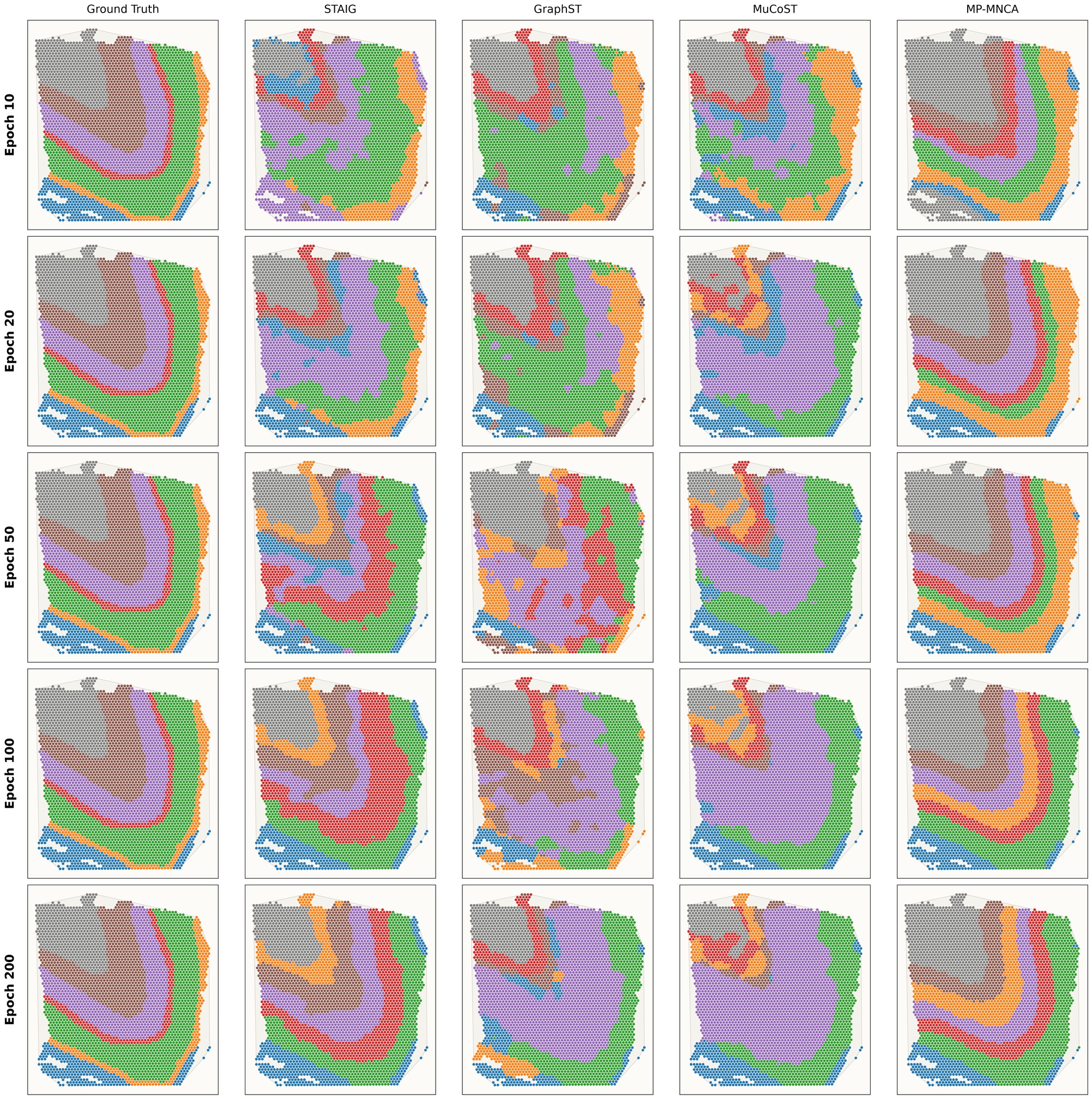}
    \caption{
    \textbf{Epoch-wise spatial-domain evolution on DLPFC section 151673.}
    The five checkpoints show how spatial-domain assignments evolve from
    early training through epoch 200, exposing differences in boundary
    consistency and regional fragmentation across methods.
    }
    \label{fig:epoch_151673}
\end{figure*}

\begin{figure*}[h]
    \centering
    \includegraphics[width=\textwidth]
    {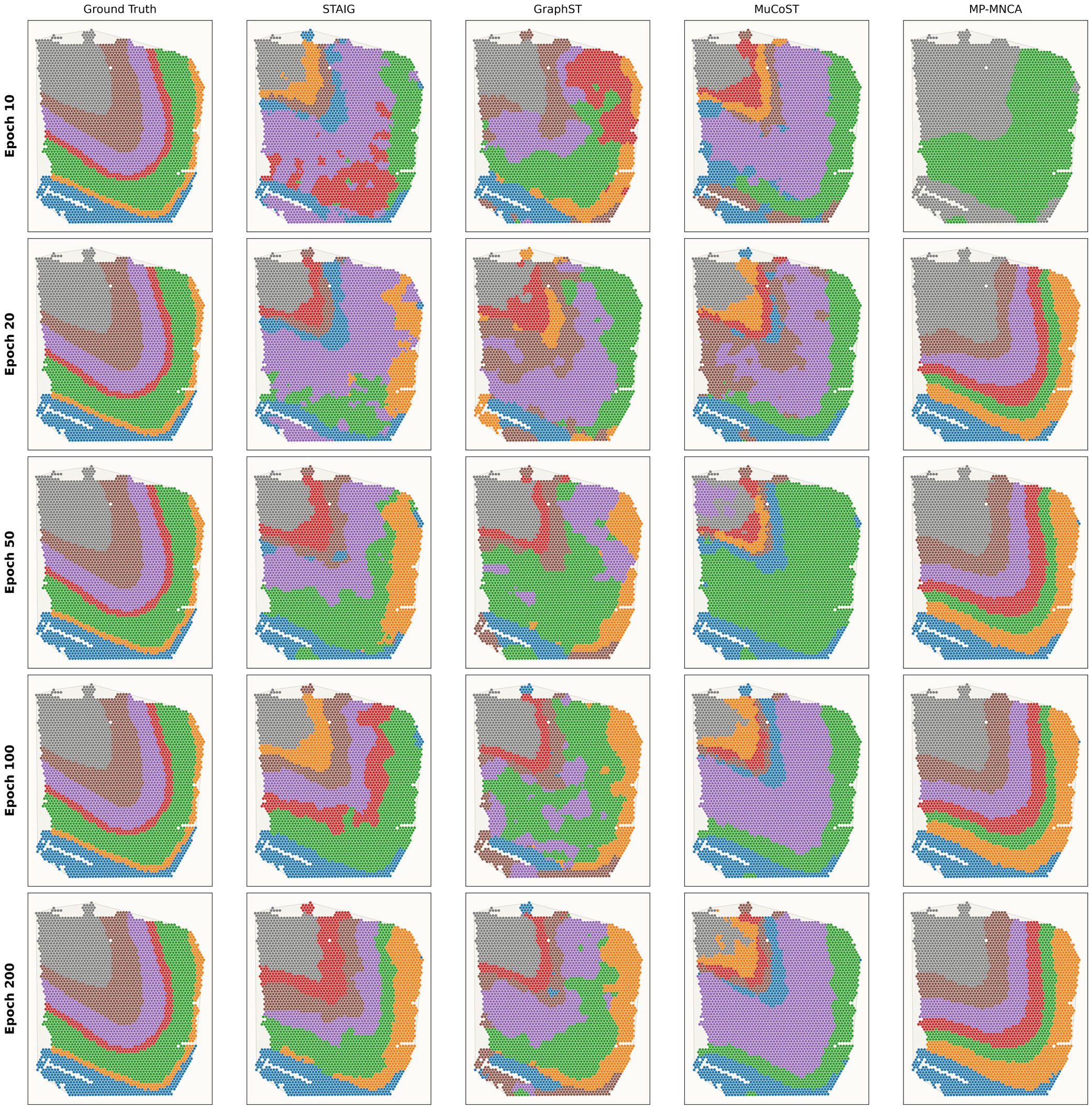}
    \caption{
    \textbf{Epoch-wise spatial-domain evolution on DLPFC section 151674.}
    The progression across checkpoints provides a qualitative comparison of
    optimization dynamics and the emergence of cortical-layer structure for
    the four evaluated methods.
    }
    \label{fig:epoch_151674}
\end{figure*}

\begin{figure*}[h]
    \centering
    \includegraphics[width=\textwidth]
    {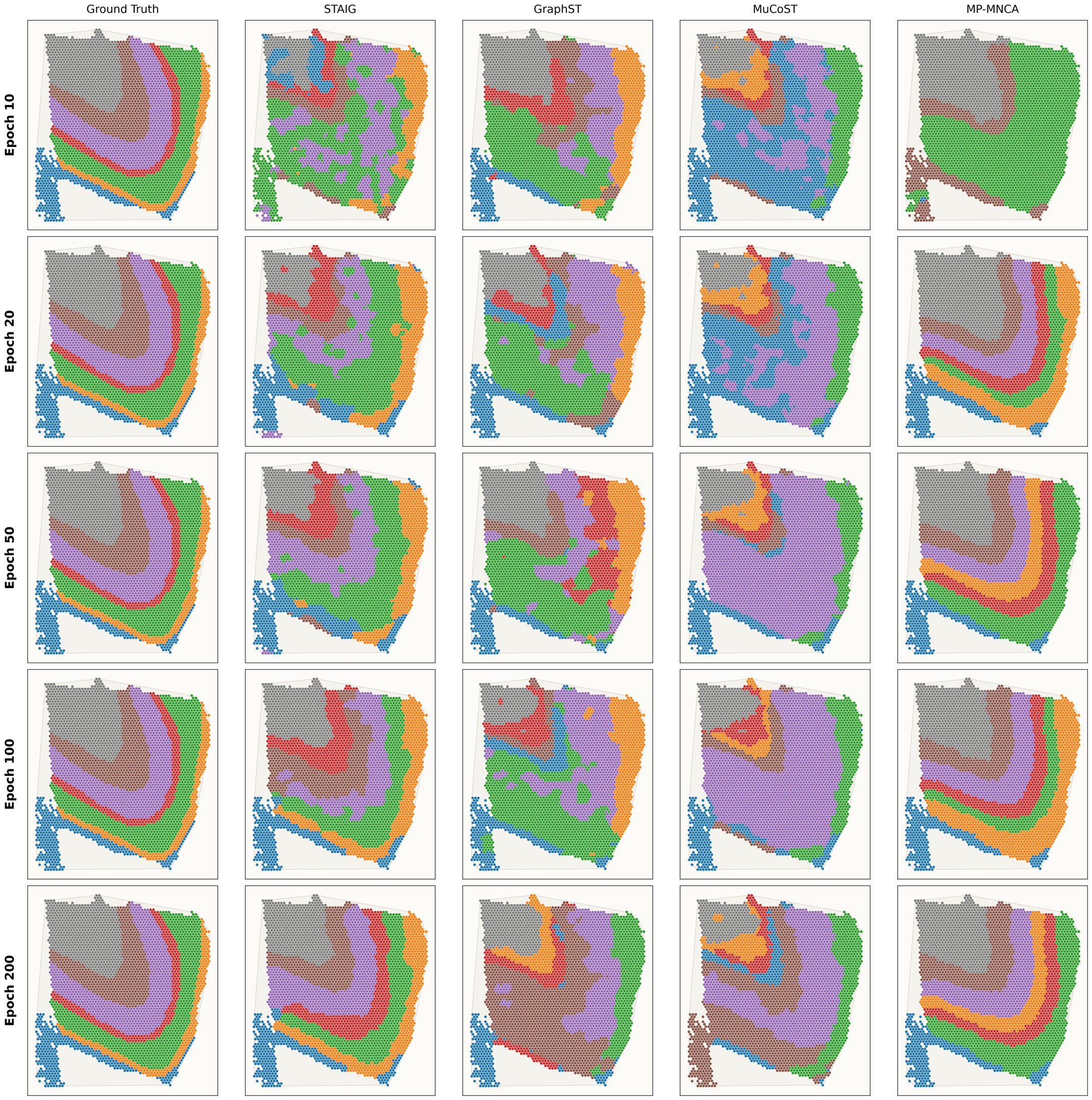}
    \caption{
    \textbf{Epoch-wise spatial-domain evolution on DLPFC section 151675.}
    This challenging section provides an important counterexample to uniform
    improvement across sections. The checkpoint maps reveal differences in
    layer fragmentation, boundary placement, and optimization behavior over
    the complete 200-epoch training budget.
    }
    \label{fig:epoch_151675}
\end{figure*}

\begin{figure*}[h]
    \centering
    \includegraphics[width=\textwidth]
    {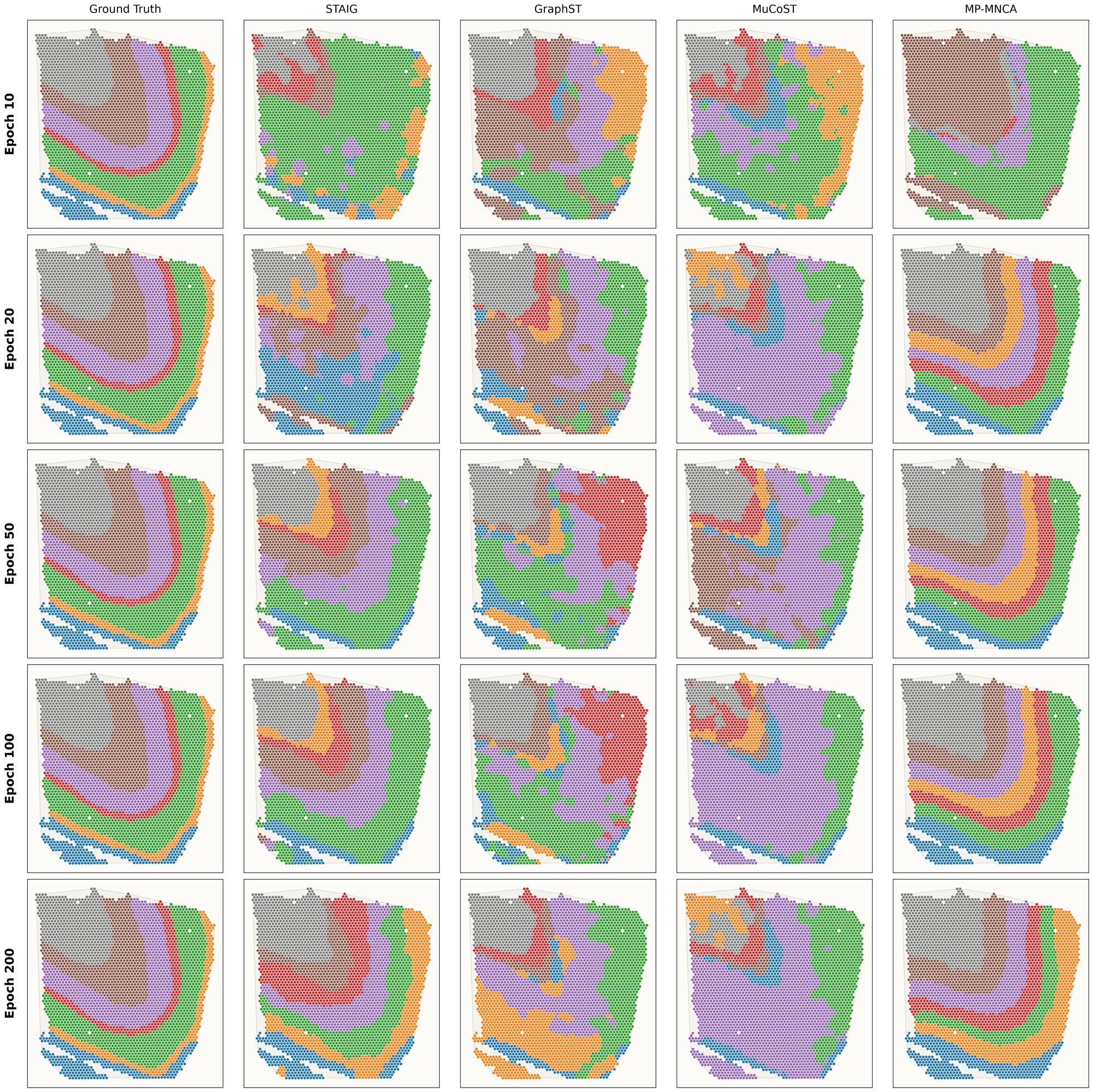}
    \caption{
    \textbf{Epoch-wise spatial-domain evolution on DLPFC section 151676.}
    Predictions at epochs 10, 20, 50, 100, and 200 complete the qualitative
    comparison across all 12 DLPFC sections and illustrate the evolution of
    cortical-domain structure under each optimization procedure.
    }
    \label{fig:epoch_151676}
\end{figure*}

\newpage